\documentclass[11pt]{article}

\usepackage[preprint]{acl}
\providecommand{\shortauthors}{}

\usepackage{times}
\usepackage{latexsym}
\usepackage[T1]{fontenc}
\usepackage[utf8]{inputenc}
\usepackage{microtype}
\usepackage{inconsolata}
\usepackage{lineno}
\usepackage{graphicx}
\usepackage{tabularx}
\usepackage[utf8]{inputenc} % allow utf-8 input
\usepackage[T1]{fontenc}    % use 8-bit T1 fonts
\usepackage{longtable, array}
\usepackage{ltablex}
\usepackage{booktabs}
\usepackage{subcaption} 
\usepackage{url}
\usepackage{graphicx}
\usepackage{amssymb}
\usepackage{amsmath}
\usepackage{amsfonts}
\usepackage{url}
\usepackage{bbm}
\usepackage{longtable}
\usepackage{rotating}
\usepackage{multirow}
\usepackage{mathrsfs}
\usepackage{enumitem}
\usepackage{adjustbox}
\usepackage{hyperref}
\usepackage{pgfplots}
\usetikzlibrary{pgfplots.dateplot}
\usepackage{filecontents}
\usepackage[many]{tcolorbox}
\usepackage{xcolor}
\usepackage{listings}
\usepackage{wrapfig}

\begin{document}

% \title{DramaDirector: Visually Aligned Generation of Short Drama Videos}
\title{DramaDirector: Geometry-Guided Short Drama Generation}

\author{
  Hengji Zhou$^{1,2*}$, Sijie Liu$^{2*}$, Jianrun Chen$^2$, Xingchen Zou$^3$,
  Lianghao Xia$^{1\dagger}$, Liqiang Nie$^1$ \\
  $^1$Harbin Institute of Technology, Shenzhen \\
  $^2$South China University of Technology \\
  $^3$The Hong Kong University of Science and Technology (Guangzhou) \\
  \texttt{hengjizhou01@gmail.com}, \texttt{cs2023lsj@mail.scut.edu.cn}, \\
  \texttt{202330450241@mail.scut.edu.cn}, \texttt{xzou428@connect.hkust-gz.edu.cn}, \\
  \texttt{aka\_xia@foxmail.com}, \texttt{nieliqiang@gmail.com}
}

\renewcommand{\shortauthors}{Zhou et al.}
\def\model{DramaDirector}
\def\dataset{DramaBoard}

\maketitle
\footnotetext[1]{$^*$Hengji Zhou and Sijie Liu have equal contribution to this work.}
\footnotetext[2]{$^\dagger$Lianghao Xia is the corresponding author.}

\begin{abstract}
Short dramas, with their rapid shot rhythms, dialogue-driven focus shifts, and demanding cinematographic grounding, pose challenges that prompt-level or text-only video generation pipelines struggle to meet. We study plot-to-short-drama generation, where a global plot and local context are transformed into visually grounded multi-shot videos. We propose \textbf{\model}, a geometry-grounded framework that lets the planner borrow cinematographic geometry from a gallery of real short-drama shots indexed by depth and pose. \model\ decouples each shot into static visual and dynamic narrative conditions, trains the planner with schema-constrained SFT and GRPO under a learned text-visual alignment reward, and retrieves depth-pose references to guide first-frame generation and image-to-video synthesis. We also introduce \textbf{\dataset}, a benchmark built from 35 live-action dramas, 2.8K episodes, and 81K shots, with structured storyboards and multi-dimensional evaluation protocols. Experiments show that \model\ improves over representative multi-agent and video generation baselines on faithfulness, consistency, and controllability. Our code is released at: \url{https://github.com/iLearn-Lab/DramaDirector}
\end{abstract}

\section{Introduction}
\label{sec:intro}

Short dramas have become a prominent form of online storytelling, marked by mobile-native distribution, fast pacing, and emotional cliffhangers~\citep{cao2026audience}. As cinematic video generation advances toward short-movie synthesis with coherent keyframes and cinematic rhythm~\citep{captaincinema,huang2025filmaster}, LLMs serve as a flexible interface for turning plots into structured narrative plans, making them a natural fit for storyboard-level short-drama generation~\citep{dsrscreen,teleki2025survey}. We focus on \emph{plot-to-short-drama generation}, which converts a global plot and local context into a multi-shot storyboard that is rendered into a short-drama video.

{
\setlength{\textfloatsep}{4pt}
\begin{figure}[t]
  \centering
  \includegraphics[width=\linewidth]{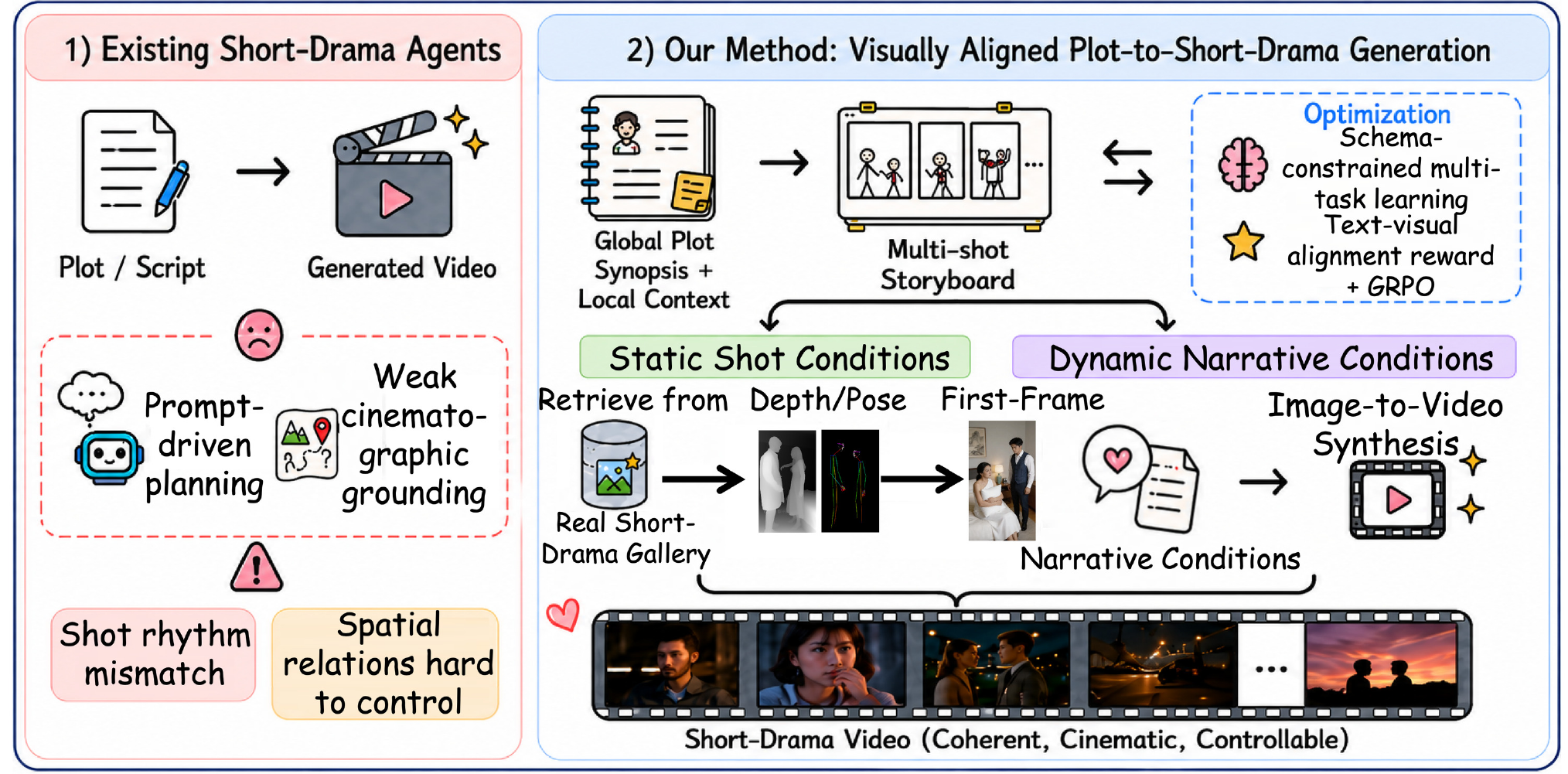}
  \vspace{-0.18in}
  \caption{Geometry-guided generation of short drama.}
  \label{fig:introduction}
  \vspace{-0.12in}
\end{figure}
}

Recent work on narrative video generation follows three technical routes. (i) \emph{Script-level supervision}, exemplified by SkyScript~\citep{skyscript100m}, uses paired scripts and shooting scripts to inject narrative priors into generation. (ii) \emph{Multi-agent orchestration}, including MovieAgent~\citep{movieagent} and GENMAC~\citep{genmac}, decomposes scenes, shots, and cinematographic decisions into chain-of-thought or collaborative agent workflows that iteratively refine compositional prompts. (iii) \emph{Controllable element composition}, such as SkyReels-A2~\citep{skyreels}, conditions video synthesis on structured visual elements to improve coherence in drama-style scenes.

For generation systems built on these planning strategies, a common deployment pipeline is to let an LLM draft a textual storyboard, use a text-to-image model to render each shot's first frame, and then animate it with an image-to-video model. This pipeline exposes two challenges that neither richer prompts nor stronger generators alone can resolve.

\textbf{A. Storyboards diverge from short-drama shot grammar.} LLMs rarely encounter large-scale short-drama corpora during pretraining and tend to write storyboards in a novelistic style~\citep{zhang2025stage,an2025onestory}, missing the fast cuts, over-the-shoulder dialogue, stacked close-ups, and one-beat-per-shot emotional rhythm that define short-drama production conventions.

\textbf{B. Text storyboards underspecify cinematographic geometry.} Even a professionally written shot description cannot convey the spatial constraints that define a frame, such as shoulder occlusion in an over-the-shoulder dialogue, the relative positions and gaze directions of speakers, or the depth ordering of bodies in a confrontation. Since these details are inherently visual, downstream generators receive only ambiguous textual cues, and reward-based alignment on text-video pairs~\citep{vader,li2025t2v} cannot recover what the text never carried.

To address these challenges, we present \model, a geometry-grounded framework that bridges the planner and the generator with real-shot visual priors. Rather than asking the LLM to verbalize cinematographic geometry, \model\ lets it \emph{borrow} geometry from a gallery of real short-drama shots pre-indexed by depth and pose skeletons, which retain compositional layout while discarding appearance noise. The planner emits a storyboard that decouples \emph{static visual conditions} from \emph{dynamic narrative conditions}: the former retrieves a depth-pose reference anchoring first-frame generation, the latter drives image-to-video synthesis. To instill the storyboard schema absent in off-the-shelf LLMs, we adapt the planner via multi-task supervised finetuning over real short-drama storyboards. We then close the loop with a geometry-aware alignment model whose retrieval similarity serves as a GRPO reward, optimizing the planner for storyboards that have real visual counterparts. In effect, \model\ turns cinematographic grounding from a text-prompting problem into a retrieval-and-reward problem.

To validate plot-to-short-drama generation, benchmarks must cover not only final video quality, but also shot-level planning and storyboard-to-video controllability. We therefore introduce \dataset, a benchmark constructed from 35 live-action dramas, covering 2.8K episodes and 81K segmented shots. Through episode-level preprocessing, transcript correction, and shot-level storyboard annotation, \dataset\ provides aligned keyframes, dialogue, episode metadata, structured storyboards, and evaluation protocols for storyboard narrative quality, storyboard-to-video instruction following, and intrinsic video quality.

Our contributions are summarized as follows:\vspace{-0.08in}
\begin{itemize}[leftmargin=*]
\item We propose \model, which generates storyboards as decoupled static and dynamic conditions, retrieves depth-pose references from real short-drama shots, and bridges structured planning with first-frame and video synthesis.\vspace{-0.08in}
\item We train the planner with multi-task schema-constrained SFT and RL via GRPO using a learned text-visual alignment reward, producing storyboards that align with real-shot references and are directly executable for video synthesis.\vspace{-0.08in}
\item We introduce \dataset, a short-drama generation benchmark with shot-level annotations, aligned keyframes, structured storyboards, and multi-dimensional evaluation protocols.\vspace{-0.08in}
\item Experiments show that \model\ improves storyboard narrative quality, storyboard-to-video controllability, and intrinsic video quality over representative baselines.
\end{itemize}
\vspace{0.05in}

\vspace{-0.1in}
\section{Methodology}
\label{sec:method}
\begin{figure*}[t]
  \centering
  \includegraphics[width=1\linewidth]{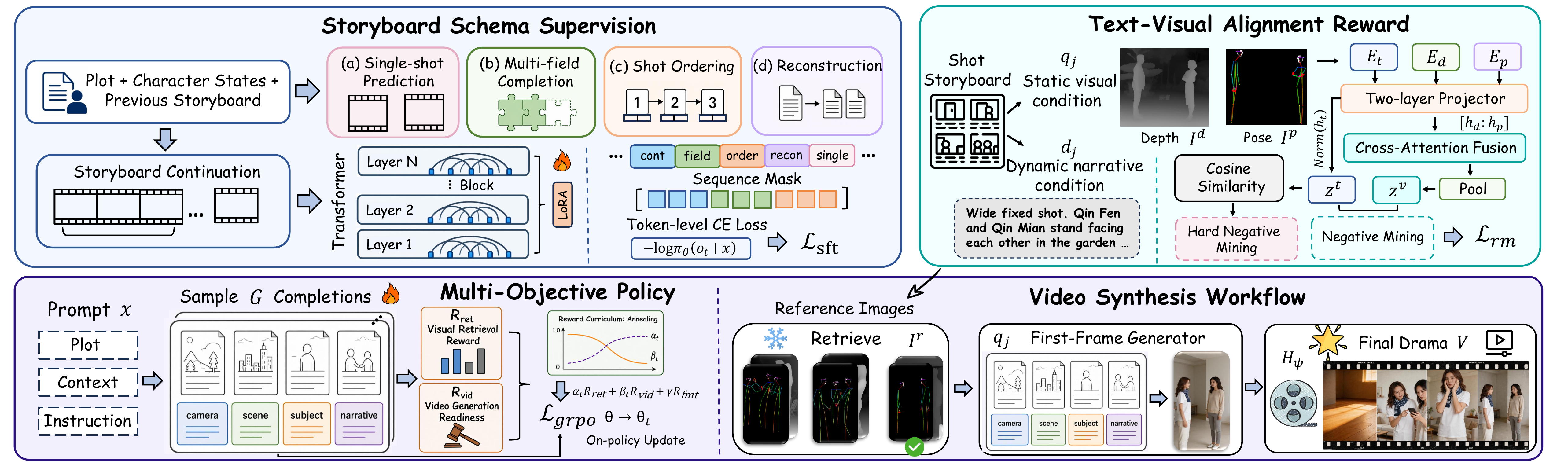}
  \vspace{-0.2in}
  \caption{Overall architecture of the proposed \model\ framework for short drama generation.} 
  \label{fig:framework}
  \vspace{-0.12in}
\end{figure*}

This section introduces our \model\ framework, whose architecture is depicted in Figure~\ref{fig:framework}.
% \vspace{-0.05in}

\subsection{Geometry-Grounded Drama Generation}
\label{sec:framework}
Given a global plot summary $\mathbf{p}\in\mathcal{P}$ and a local narrative context $\mathbf{c}\in\mathcal{C}$, short drama generation aims to produce a multi-shot video $\mathcal{V}=\{V_j\}_{j=1}^{N}$ faithful on two fronts: \emph{narrative faithfulness} to $(\mathbf{p},\mathbf{c})$, naturally expressed in text, and \emph{geometric faithfulness} in each shot's spatial layout, body configuration, and depth relations, requiring spatial rather than linguistic cues. Direct text-to-video synthesis thus systematically underspecifies the latter, motivating a structured storyboard intermediate $\mathcal{Y}$ that decomposes the task into planning and realization stages:
\begin{align}
    g_\phi:(\mathbf{p},\mathbf{c})\mapsto\mathcal{Y},
\end{align}
where $\mathcal{Y}=\{(q_j, d_j)\}_{j=1}^{N}$ is a sequence of shot-level entries, each pairing a \emph{static visual condition} $q_j$ (camera framing, scene) anchoring the keyframe with a \emph{dynamic narrative condition} $d_j$ (action, dialogue, emotion) unfolding within the shot.

Yet $q_j$ remains textual and leaves the underlying geometry unspecified. We therefore ground each shot in a real-world geometric prior by retrieving a depth--pose reference $I^r_j$ from a pre-indexed gallery $\mathcal{G}$ of short-drama shots (see Appendix~\ref{sec:benchmark}), from which the keyframe $I^0_j$ is then synthesized:
\begin{align}
    I^r_j = \arg\max_{I_i\in\mathcal{G}}\;(\mathbf{z}^t_j)^\top \mathbf{z}^v_i,~~
    I^0_j = F_\xi(q_j,\,I^r_j),
\end{align}
where $\mathbf{z}^t_j$ and $\mathbf{z}^v_i$ are text- and visual-side embeddings, whose inner product measures the geometric compatibility between $q_j$ and each gallery shot, and $F_\xi$ is a first-frame generator that renders keyframe $I^0_j$ by conditioning on $q_j$ while inheriting layout, pose, and depth from $I^r_j$.

With the keyframe, an image-to-video model $H_\psi$ rolls out the shot using $I^0_j$ as the visual anchor and $d_j$ as the temporal instruction, and the final drama is obtained by concatenating all shots:
\begin{align}
V_j = H_\psi(I^0_j,\, d_j),~~ \mathcal{V}=(V_1,\ldots,V_N),
\end{align}

\subsection{Storyboard Supervised Finetuning}
\label{sec:sft}
The framework above hinges on the storyboard planner $g_\phi$, yet off-the-shelf LLMs are unaware of the short-drama storyboard schema, including its camera grammar, field structure, and shot-to-shot transitions. We therefore adapt $g_\phi$ via supervised finetuning on real short-drama storyboards.

\subsubsection{Multi-Task Storyboard Supervision}
Since no single training signal can capture all aspects of storyboard quality, we decompose our SFT task into one main objective and four auxiliary objectives, $\mathcal{T}_{\mathrm{sft}}=\{\tau_{\mathrm{cont}},\tau_{\mathrm{single}},\tau_{\mathrm{field}},\tau_{\mathrm{order}},\tau_{\mathrm{recon}}\}$, each objective detailed as below.

Each annotated episode $s=(\mathbf{p},\mathbf{c},\mathcal{Y})$ with $\mathcal{Y}=\{(q_j,d_j)\}_{j=1}^{N}$ is obtained by automatic shot segmentation, keyframe--dialogue alignment, and schema-conditioned VLM storyboard annotation. From each sample, a task-specific renderer $\mathcal{A}_{\tau}$ produces a prompt--target pair:
\begin{equation}
(x_{\tau},\mathcal{Y}_{\tau})=\mathcal{A}_{\tau}(s),
\quad
\mathcal{D}_{\tau}=\{(x_{\tau},\mathcal{Y}_{\tau})\}_{s\in\mathcal{D}_{0}},
\end{equation}
where $\mathcal{D}_{0}$ is the structured storyboard corpus.

\noindent\textbf{Storyboard Continuation.} Given $\mathbf{p}$, $\mathbf{c}$, and a prefix of past shots $q_{<j}$, the main task $\tau_{\mathrm{cont}}$ autoregressively predicts the next several shots $q_{j:j+k}$, directly mirroring the inference-time use of $g_\phi$.

\noindent\textbf{Auxiliary Objectives.} (i)~\textbf{Single-shot Prediction} predicts the next shot from local context to improve local realization; (ii)~\textbf{Multi-field Completion} recovers missing fields from partial specifications to capture cross-field dependencies; (iii)~\textbf{Shot Ordering} reorders shuffled sequences to strengthen temporal coherence; and (iv)~\textbf{Summary-guided Reconstruction} reconstructs storyboards from summaries to reinforce plan-to-shot grounding. Together, these tasks provide a stable initialization for the reinforcement learning stage.\vspace{-0.05in}

\subsubsection{Training Objective}
We optimize the base model with LoRA adapters on the attention and feed-forward projections. For each training sample, let $\mathcal{O}=(o_1,\ldots,o_T)$ be the tokenized target storyboard and $m_t$ indicate response tokens after the prompt boundary. The objective can be formalized as follows:
{\setlength{\abovedisplayskip}{5pt}\setlength{\belowdisplayskip}{5pt}
\begin{equation}
\mathcal{L}_{\mathrm{sft}}
=-\sum_{t=1}^{T}m_t
\log\pi_{\theta_{\mathrm{llm}}}(o_t\mid x,o_{<t}),
\end{equation}}

\noindent where $x$ is the task prompt and $o_{<t}$ denotes previous target tokens. For short auxiliary tasks, multiple samples are packed into one training sequence under the maximum length constraint, with attention restricted within each sample's boundary so that the loss is computed independently per sample.
% \vspace{-0.1in}

\subsection{RL from Retrieval and Rendering}
\label{sec:rl}
To align $g_\phi$ with retrieval from $\mathcal{G}$ and rendering by $H_\psi$, we continue training the SFT-initialized planner with reinforcement learning. We adopt Group Relative Policy Optimization (GRPO), whose value-network-free advantage estimation fits the long, structured nature of storyboard outputs.

\noindent\textbf{Visual Retrieval Reward.}
For generated shots $\{y_j\}$ paired with target episode $e$, each static condition $q_j$ is mapped to its text-side alignment vector $\mathbf{z}^{t}_{j}$ by the alignment model introduced in Section~\ref{sec:alignment}. Let $\mathcal{Z}^{\pm}_{e}$ denote the visual-vector sets from the target and sampled negative episodes. We compute top-$k$ retrieval similarities on each side:
{\setlength{\abovedisplayskip}{7pt}\setlength{\belowdisplayskip}{7pt}
\begin{equation}
\mathcal{S}^{\pm}_{j}=\operatorname{TopK}\{(\mathbf{z}^{t}_{j})^\top\mathbf{z}^v:\mathbf{z}^v\in\mathcal{Z}^{\pm}_{e}\},
\end{equation}}

\noindent where $\operatorname{TopK}$ returns the $k$ largest scalar similarities. Shot- and completion-level scores are then aggregated into a contrastive reward:
{\setlength{\abovedisplayskip}{7pt}\setlength{\belowdisplayskip}{7pt}
\begin{align}
&R_{\mathrm{ret}}~=~\frac{1}{2}~+~\frac{\bar{c}^{+}-\bar{c}^{-}}{4},\nonumber\\
c_j^{\pm}&=\frac{1}{k}\sum_{a\in\mathcal{S}^{\pm}_{j}}a,~~~
\bar{c}^{\pm}=\frac{1}{N}\sum_{j=1}^{N} c_j^{\pm},
\end{align}}

\noindent This contrastive form rewards storyboards matching the target episode while discouraging generic visual descriptions matching unrelated episodes.
\\\vspace{-0.12in}

\noindent\textbf{Video-Generation Readiness Reward.}
Beyond static grounding, video synthesis requires dynamic conditions that are concrete and temporally coherent. We evaluate the sequence $\{d_j\}_{j=1}^{N}$ with an LLM judge $J$, formally defined as:
{\setlength{\abovedisplayskip}{7pt}\setlength{\belowdisplayskip}{7pt}
\begin{equation}
R_{\mathrm{vid}}=J(\{d_j\}_{j=1}^{N})\in[0,1],
\end{equation}}

\noindent where $J$ scores action concreteness, visual detail, transition coherence, emotional progression, and short-drama pacing over the full sequence.

\noindent\textbf{Reward Aggregation.}
We further introduce a lightweight schema-validity reward $R_{\mathrm{fmt}}\in[0,1]$ that penalizes missing fields and unparsable outputs, and combine the three terms as:
{\setlength{\abovedisplayskip}{7pt}\setlength{\belowdisplayskip}{7pt}
\begin{align}
R(\mathcal{Y})&=\alpha_s R_{\mathrm{ret}}+\beta_s R_{\mathrm{vid}}+\gamma R_{\mathrm{fmt}},
\nonumber\\
&\alpha_s=\alpha_0+(\alpha_T-\alpha_0)\,p_s,
\end{align}}

\noindent where $s$ is the current update step, $p_s=\min(s/T,1)$, $T$ is the annealing horizon, and $\alpha_0,\alpha_T$ ($\beta_0,\beta_T$) are the start/end weights of the linearly annealed $\alpha_s$ ($\beta_s$); $\gamma$ stays fixed. The curriculum lets the policy first secure schema validity and visual grounding, then gradually shift weight toward the harder video-readiness objective.

\noindent\textbf{Group Relative Policy Optimization.}
Given a prompt $x$, we sample $G$ completions and optimize the clipped objective as follows:
{\setlength{\abovedisplayskip}{7pt}\setlength{\belowdisplayskip}{7pt}
\begin{equation}
\mathcal{L}_{\mathrm{rl}}=-\frac{1}{G}\sum_{i=1}^{G}\frac{1}{T_i}\sum_{t=1}^{T_i}\min(r_{i,t}A_i,\bar{r}_{i,t}A_i),
\end{equation}}

\noindent where $T_i$ is the response length and $A_i$ is obtained by normalizing $R(\mathcal{Y}_i)$ across the $G$ completions. The ratio $r_{i,t}$ compares the current policy with the sampling policy at token $t$, and $\bar{r}_{i,t}$ clips it to $[1-\epsilon,1+\epsilon]$ with clipping coefficient $\epsilon$.

\subsection{Geometry-Aware Text-Visual Alignment}
\label{sec:alignment}
Both the geometry-grounded retrieval in Section~\ref{sec:framework} and the retrieval reward in Section~\ref{sec:rl} reduce to one primitive: scoring how well a static condition matches the geometry of a candidate shot. Off-the-shelf vision-language encoders, trained on raw RGB, entangle appearance with geometry and tend to favor semantically similar yet geometrically mismatched shots. We thus train a dedicated alignment model on static conditions and depth--pose evidence, and freeze it as the shared backbone for inference-time retrieval and policy-time reward.

\noindent\textbf{Textual Condition Decomposition.}
Each storyboard includes a static visual condition and a dynamic narrative condition. The static one concatenates the camera, scene, and subject-state fields,
{\setlength{\abovedisplayskip}{6pt}\setlength{\belowdisplayskip}{6pt}
\begin{equation}
q_j = [a_j; e_j; u_j],
\end{equation}}

\noindent serving as the prompt for first-frame generation, while the dynamic $d_j$ from the narrative field drives video synthesis. Only $q_j$ participates in alignment, since geometric grounding is a property of the keyframe rather than the temporal rollout.

\noindent\textbf{Architecture.}
For shot $y_j$, the visual evidence $I_j=(I_j^{d},I_j^{p})$ consists of the depth map and the human-pose map of the keyframe, deliberately stripping appearance and identity to retain only geometric cues. Modality embeddings are obtained by encoders $E_t$, $E_d$, $E_p$ as follows:
{\setlength{\abovedisplayskip}{7pt}\setlength{\belowdisplayskip}{7pt}
\begin{equation}
\mathbf{x}^{t}_{j}=E_t(q_j),~~
\mathbf{x}^{d}_{j}=E_d(I_j^{d}),~~
\mathbf{x}^{p}_{j}=E_p(I_j^{p}),
\end{equation}}

\noindent where $\mathbf{x}^{t}_{j},\mathbf{x}^{d}_{j},\mathbf{x}^{p}_{j}\in\mathbb{R}^{D}$. Modality-specific two-layer projection blocks (with normalization, GELU, and dropout) map them to $D_a$ dimensions:
{\setlength{\abovedisplayskip}{6pt}\setlength{\belowdisplayskip}{6pt}
\begin{equation}
\mathbf{h}^{r}_{j}=P_r(\mathbf{x}^{r}_{j}),\quad r\in\{t,d,p\}.
\end{equation}}

\noindent The text projection is $\ell_2$-normalized to form the text-side vector, while the two visual projections are stacked as modality tokens and fused by the self-attention mechanism:
{\setlength{\abovedisplayskip}{7pt}\setlength{\belowdisplayskip}{7pt}
\begin{align}
\mathbf{z}^{t}_{j}&=\operatorname{norm}(\mathbf{h}^{t}_{j}),~~
\mathbf{M}_{j}=[\mathbf{h}^{d}_{j};\mathbf{h}^{p}_{j}],\nonumber\\
\widetilde{\mathbf{M}}_{j}&=\operatorname{LN}(\mathbf{M}_{j}+\operatorname{MHA}(\mathbf{M}_{j})),\nonumber\\
\mathbf{z}^{v}_{j}&=\operatorname{norm}(W_v\operatorname{Pool}(\widetilde{\mathbf{M}}_{j})+\mathbf{b}_v),
\end{align}}

\noindent where $\operatorname{Pool}$ averages the two modality tokens, and $W_v,\mathbf{b}_v$ are the visual output projection parameters. The resulting $\mathbf{z}^{t}_{j}$ and $\mathbf{z}^{v}_{j}$ are exactly the vectors consumed by the gallery retriever in Section~\ref{sec:framework} and by the retrieval reward $R_{\mathrm{ret}}$ in Section~\ref{sec:rl}.

\noindent\textbf{Contrastive Learning.}
We train the alignment model with bidirectional contrastive learning. Positive pairs are the text and visual vectors of the same shot; negatives are drawn from non-adjacent shots, different episodes, and different dramas to cover progressively coarser mismatches. Given anchors $\{\mathbf{a}_i\}_{i=1}^{B}$, positives $\{\mathbf{b}_i\}_{i=1}^{B}$, and negative sets $\{\mathcal{N}_i\}_{i=1}^{B}$, the contrastive learning is defined as:
{\setlength{\abovedisplayskip}{6pt}\setlength{\belowdisplayskip}{6pt}
\begin{align}
\kappa_i(\mathbf{b})&=\exp(cos(\mathbf{a}_i,\mathbf{b})/\tau),\\
\ell(\mathbf{a},\mathbf{b},\mathcal{N})
=-\frac{1}{B}&\sum_{i=1}^{B}\log
\frac{\kappa_i(\mathbf{b}_i)}{\kappa_i(\mathbf{b}_i)+\sum_{\mathbf{c}\in\mathcal{N}_i}\kappa_i(\mathbf{c})},\nonumber
\end{align}}

\noindent where $B$ is the batch size and $\tau$ the temperature. Letting $\mathcal{N}^{v}_i$ and $\mathcal{N}^{t}_i$ denote $K$ sampled visual and textual negatives for anchor $i$, the bidirectional InfoNCE loss is defined as follows:
{\setlength{\abovedisplayskip}{6pt}\setlength{\belowdisplayskip}{6pt}
\begin{equation}
\mathcal{L}_{\mathrm{nce}}
=\tfrac{1}{2}\big[
\ell(\mathbf{z}^{t},\mathbf{z}^{v+},\mathcal{N}^{v})
+\ell(\mathbf{z}^{v},\mathbf{z}^{t+},\mathcal{N}^{t})
\big].
\end{equation}}

\noindent Because shots within the same drama share recurring scenes and characters, random negatives quickly become trivial and yield only coarse alignment. We therefore mine hard visual negatives from within each anchor's own drama:
{\setlength{\abovedisplayskip}{6pt}\setlength{\belowdisplayskip}{6pt}
\begin{equation}
\mathcal{N}^{v,h}_i=\operatorname{TopK}_{K_h}\{\mathbf{z}^{v}_j:j\in\mathcal{C}^{\mathrm{dr}}_i,~s_{ij}<\eta_i\},
\end{equation}}

\noindent where $\mathcal{C}^{\mathrm{dr}}_i$ is the same-drama candidate set, $s_{ij}$ is cosine similarity, and $\eta_i$ excludes the top 10\% most similar shots to guard against false negatives. The hard-negative loss can be defined as:
{\setlength{\abovedisplayskip}{6pt}\setlength{\belowdisplayskip}{6pt}
\begin{equation}
\mathcal{L}_{\mathrm{hard}}=\ell(\mathbf{z}^{t},\mathbf{z}^{v+},\mathcal{N}^{v,h})
\end{equation}}

\noindent is combined with the bidirectional term to form the final objective, as follows:
{\setlength{\abovedisplayskip}{6pt}\setlength{\belowdisplayskip}{6pt}
\begin{equation}
\mathcal{L}_{\mathrm{rm}}=\lambda_1\mathcal{L}_{\mathrm{nce}}+\lambda_2\mathcal{L}_{\mathrm{hard}}.
\end{equation}}

\noindent Once trained, the alignment model is frozen and supplies the geometric similarity signal for both retrieval and reward throughout the framework.

\section{Evaluation}
\begin{table}
    \small
    \centering
    \caption{Statistics of the experimental datasets.}
    \label{tab:dataset}
    \setlength{\tabcolsep}{0.7mm}
    \vspace{-0.12in}
    \begin{tabular}{ccccccc}
      \hline
      Task & Drama & Episodes & Shots & Dur & Samples\\
      \hline
      Continuation & 35 & 2,807 & 81,283 & 50.2h & 13,262\\
      Shot prediction & 35 & 2,812 & 27,437 & 16.9h & 27,437\\
      Field completion & 35 & 2,455 & 6,660 & 4.1h & 6,660\\
      Shot ordering & 35 & 2,437 & 30,534 & 18.7h & 6,893\\
      Reconstruction & 35 & 2,767 & 58,485 & 35.9h & 15,121\\
      \hline
    \end{tabular}
    \vspace{-0.15in}
\end{table}

We evaluate \model\ across six questions: \textbf{RQ1} end-to-end short drama generation against representative baselines; \textbf{RQ2} analyzes key modules; \textbf{RQ3} retrieval discriminability of the text-visual alignment model; \textbf{RQ4} quality of retrieval-grounded first-frame generation; \textbf{RQ5} studies hyperparameter sensitivity; \textbf{RQ6} qualitatively demonstrates the effectiveness of \model\ .\vspace{-0.05in}

\subsection{Experimental Settings}
\label{sec:exp_setting}
\subsubsection{Datasets and Evaluation Protocols}
\dataset\ is constructed from 35 live-action short dramas segmented into shot-level timelines with aligned keyframes and dialogue (Table~\ref{tab:dataset}). All auxiliary tasks, the text-visual alignment model, and downstream generation experiments follow the train/validation/test split of the storyboard continuation corpus at an 8:1:1 ratio. We randomly sample 300 videos and 600 images from the test set~\citep{visioncreator}. Judge-scored criteria use a 3-point scale $\{0,0.5,1\}$, while Similarity and DIS are reported as automatic continuous metrics. Evaluation covers three dimensions:
\textbf{(i) Storyboard narrative quality}~\citep{chhun2022human} -- \textit{Relevance} (alignment with context), \textit{Coherence} (logical transitions and pacing), and \textit{Engagement} (emotional resonance and rhythm);
\textbf{(ii) Instruction following}~\citep{feng2026gen} -- \textit{Faithfulness} (conformance to directives), \textit{Similarity} (cosine similarity between Qwen3-VL embeddings of generated videos and dynamic narrative prompts), and \textit{Purity} (absence of hallucinations);
\textbf{(iii) Video intrinsic quality}~\citep{guo2026comfymind} -- \textit{Aesthetics} (lighting, color grading, and composition), \textit{Consistency} (stable appearance and geometry), and \textit{DIS} (masked SSIM distance after warping adjacent 1-FPS frames with DIS optical flow; lower is better).
For more details, please refer to Appendix~\ref{sec:benchmark} and Appendix~\ref{sec:judge_prompt}.\vspace{-0.05in}

%\textit{Correctness} (accuracy of subjects, props, and layout),

\begin{table*}[t]
\centering
\setlength{\tabcolsep}{0.6mm}
% Adjust table row spacing for readability.
\renewcommand{\arraystretch}{1.1} 
\caption{Overall performance comparison across eight baseline settings. Best and second-best results are highlighted in bold and underlined, excluding the two variants. Our standard deviations are computed over five runs.}
\label{tab:overall}
\small
\vspace{-0.12in}
\begin{tabular}{c|ccc|ccc|ccc}
\hline
\multirow{2}{*}{Model} & \multicolumn{3}{c|}{Storyboard Narrative Quality} & \multicolumn{3}{c|}{Instruction Following} & \multicolumn{3}{c}{ Video Intrinsic Quality} \\
\cline{2-10}
 & Relevance & Coherence & Engagement & Faithfulness & Purity & Similarity & Aesthetics & Consistency & DIS $\downarrow$ \\
\hline
SkyScript & 0.4850 & 0.4900 & 0.4850 & 0.6667 & 0.8333 & 0.5855 & 0.6467 & 0.7000 & 0.1753 \\
ShoulderShot & 0.5800 & 0.6233 & 0.5933 & 0.6933 & 0.9333 & 0.5275 & 0.6917 & 0.6933 & 0.1618 \\
SkyReels-A2 & 0.5350 & 0.5333 & 0.5150 & 0.7517 & 0.9233 & 0.5626 & 0.5900 & 0.6600 & \underline{0.1596} \\
Dreamrunner & 0.5850 & 0.5850 & 0.5683 & \textbf{0.7950} & \underline{0.9667} & 0.6098 & 0.6067 & \textbf{0.9283} & 0.1942 \\
\hline
GenMac & 0.5300 & 0.5300 & 0.5067 & 0.7150 & 0.8800 & 0.6267 & 0.6033 & 0.7483 & 0.2026 \\
MovieAgent & 0.6000 & 0.6000 & 0.5550 & 0.7100 & 0.8733 & 0.6340 & 0.6283 & 0.7367 & 0.1903 \\
UniVA & \underline{0.6633} & \underline{0.6800} & \textbf{0.6200} & 0.6200 & 0.8900 & 0.6028 & \underline{0.7167} & 0.7733 & 0.1799 \\
VideoAuteur & 0.5900 & 0.5950 & 0.5783 & 0.7450 & 0.9533 & \underline{0.6629} & 0.6350 & 0.6717 & 0.1657 \\
\hline
% \model & \textbf{0.7150} & \textbf{0.7750} & \underline{0.6017} & \textbf{0.8333} & \underline{0.7208} & \textbf{0.9712} & \underline{0.7433} & \textbf{0.9425} & \textbf{0.6783} \\
 \model & \textbf{0.7230} & \textbf{0.7732} & \underline{0.6032} & \underline{0.7712} & \textbf{0.9752} & \textbf{0.6826} & \textbf{0.8189} & \underline{0.9003} & \textbf{0.1549} \\
+ Wan 2.6 & 0.7817 & 0.7650 & 0.6633 & 0.7217 & 0.8850 & 0.6643 & 0.7183 & 0.8400 & 0.1881 \\
+ NanoBanana 2 & 0.6800 & 0.6933 & 0.6167 & 0.7700 & 0.8967 & 0.6526 & 0.5750 & 0.7617 & 0.1663\\
\hline
$\sigma$ & $\pm 0.0253$ & $\pm 0.0202$ & $\pm 0.0141$ & $\pm 0.0123$ & $\pm 0.0086$ & - & $\pm 0.0166$ & $\pm 0.0130$ & - \\
\hline

\end{tabular}
\vspace{-0.05in}
\end{table*}

\subsubsection{Baseline Methods}
\model\ is compared with a comprehensive set of baselines and adaptations, including \textbf{(i) Short-Drama Generation Methods}: SkyScript~\citep{skyscript100m}, ShoulderShot~\citep{shouldershot}, SkyReels-A2~\citep{skyreels}, and Dreamrunner~\citep{dreamrunner}; \textbf{(ii) Image Generation Frameworks}: ComfyAgent~\citep{comfybench}, ComfyMind~\citep{guo2026comfymind}, and Gen-Search~\citep{feng2026gen}; \textbf{(iii) Generative Agentic Frameworks}: GENMAC~\citep{genmac}, MovieAgent~\citep{movieagent}, UniVA~\citep{univa}, and VideoAuteur~\citep{videoauteur}---for a rigorous comparison, these agentic baselines are deliberately powered by Qwen3.5-Flash, which possesses stronger inherent capabilities than the lightweight 8B planner used in \model\, to ensure maximum planning capability; and \textbf{(iv) Retrieval-Augmented Frameworks}: LightRAG~\citep{lightrag}, RAG-Anything~\citep{raganything}, and VidoRAG~\citep{pathorag}.

\subsubsection{Implementation Details}
All methods are implemented following their original specifications. To ensure a fair comparison of planning and grounding strategies, all baselines and \model\ use the same downstream generation backbones: Seedream-v5-lite for image generation and Vidu-q3-turbo for video synthesis, except for the explicitly marked generator-swap variants. \model\ adopts Qwen3-8B as the base model. For text-visual alignment training, $\lambda_1=\lambda_2=2.0$, learning rate $1 \times 10^{-2}$, with 50 negatives and 54 hard negatives, and early stopping at step~3. For schema-constrained supervision, lora rank is set to 16 with learning rate $5 \times 10^{-6}$ and AdamW optimizer. For multi-objective policy optimization, group size is 4, maximum completion length is 2048, and learning rate is $5 \times 10^{-7}$; the video-readiness reward is judged by Qwen3.5-Flash, while the final evaluation uses Qwen3.5-Plus. Both supervision stages share warm-up ratio 0.01, weight decay $1 \times 10^{-3}$ and 3 training epochs. We employ tongyi-embedding-vision-flash-2026-03-06 for multimodal encoding.\vspace{-0.05in}

\subsection{Overall Performance Comparison (RQ1)}

% We compare \model\ against eight representative baselines across all nine metrics in Table~\ref{tab:overall}. \textbf{For Storyboard Narrative Quality}, \model\ achieves the strongest overall results and leads Relevance and Coherence, showing that multi-task schema supervision and multi-objective reward optimization better capture continuity and short-drama pacing. \textbf{For Instruction Following}, \model\ obtains the highest Correctness and Purity and ranks second on Faithfulness, indicating that schema-constrained planning and retrieval-grounded synthesis improve camera, composition, and object control while reducing hallucinations. \textbf{For Video Intrinsic Quality}, \model\ ranks first on Aesthetics and Motion and second on Consistency, suggesting that depth--pose grounded references stabilize appearance, scene geometry, and motion. Although \model\ is surpassed by UniVA on Engagement and by Dreamrunner on Faithfulness and Consistency, its overall gains over prompt-driven agentic baselines powered by Qwen3.5-Flash highlight the value of our short-drama-aware planning and visual grounding.\vspace{-0.05in}
We compare \model\ against eight baselines in Table~\ref{tab:overall}. Excluding generator-swap variants, \model\ leads Relevance and Coherence on storyboard narrative quality, showing that multi-task schema supervision and reward optimization better capture continuity and short-drama pacing. For \textbf{Instruction Following}, \model\ obtains the highest Similarity and Purity, indicating stronger alignment with dynamic narrative prompts and fewer unprompted elements. For \textbf{Video Intrinsic Quality}, \model\ achieves the highest Aesthetics and the lowest DIS, while ranking second on Consistency, suggesting improved visual fidelity and temporal stability. Although UniVA is slightly better on Engagement and Dreamrunner leads Faithfulness and Consistency, \model's gains over the Qwen3.5-Flash-powered agentic baselines highlight the value of short-drama-aware planning. The supplementary variants further indicate that the planner--retrieval interface can be paired with different generation backbones, although performance varies with each backend's control fidelity. We provide additional human evaluation details in Appendix~\ref{sec:human_eval}.\vspace{-0.05in}

\begin{table}
  \centering
  \small
  \setlength{\tabcolsep}{0.6mm}
  \renewcommand{\arraystretch}{1.1} % Improve table readability.
  \caption{Performance of ablated \model.}
  \label{tab:ablation}
  \vspace{-0.12in}
  \begin{tabular}{ c|c| cc|cc }
  \hline
\multicolumn{2}{c|}{Metric} & Faithfulness & Purity & Consistency & DIS $\downarrow$\\
\hline
  \multirow{4}{*}{\rotatebox{90}{\textbf{Generation}}} & Text-only & 0.6083 & 0.9250 & 0.5750 & \underline{0.1847} \\
  \cline{2-6}
  & Depth-only & \underline{0.6583} & 0.8667 & \underline{0.6333} & 0.2065 \\
  \cline{2-6}
  & Pose-only & 0.6333 & \underline{0.9500} & 0.5750 & 0.2143 \\
  \cline{2-6}
  & \textbf{Origin} & \textbf{0.7712} & \textbf{0.9752} & \textbf{0.9003} & \textbf{0.1549} \\
  \hline
  \hline
  \multicolumn{2}{c|}{Metric} & R@10 & N@10 & R@20 & N@20 \\
  \hline
  \multirow{5}{*}{\rotatebox{90}{\textbf{Retrieval}}} &  w/o $\mathcal{L}_{\mathrm{nce}}$ & 0.0752 & 0.0368 & 0.1198 &  0.0480 \\
  \cline{2-6}
  & w/o $\mathcal{L}_{\mathrm{hard}}$ & \underline{0.0836} & \underline{0.0416} & \underline{0.1323} & \underline{0.0538} \\
  \cline{2-6}
  & w/o pose & 0.0184 & 0.0089 & 0.0343 & 0.0129 \\
  \cline{2-6}
  & w/o depth & 0.0013 & 0.0004 & 0.0031 & 0.0009 \\
  \cline{2-6}
  & \textbf{Origin} & \textbf{0.0954} & \textbf{0.0483} & \textbf{0.1429} & \textbf{0.0603} \\
  \hline  
  \end{tabular}
  \vspace{-0.2in}
\end{table}

\subsection{Ablation Study (RQ2)}
\begin{figure}[t]
  \centering
  
  % --- Legend ---
  \includegraphics[width=\linewidth]{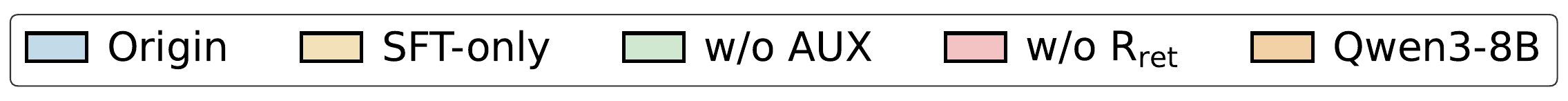}
  \vspace{-0.15in}

  % --- Image-text group ---
  \begin{minipage}[b]{0.32\linewidth}
    \centering
    \includegraphics[width=\linewidth]{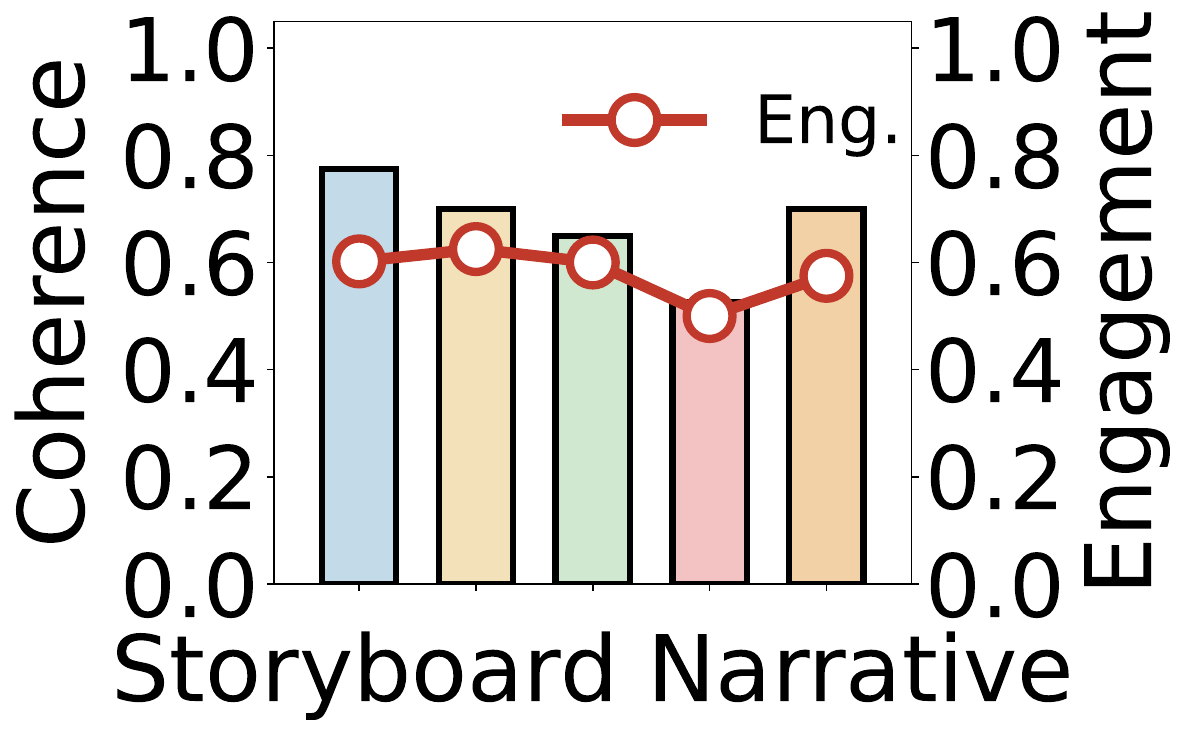}
  \end{minipage}
  \hfill
  \begin{minipage}[b]{0.32\linewidth}
    \centering
    \includegraphics[width=\linewidth]{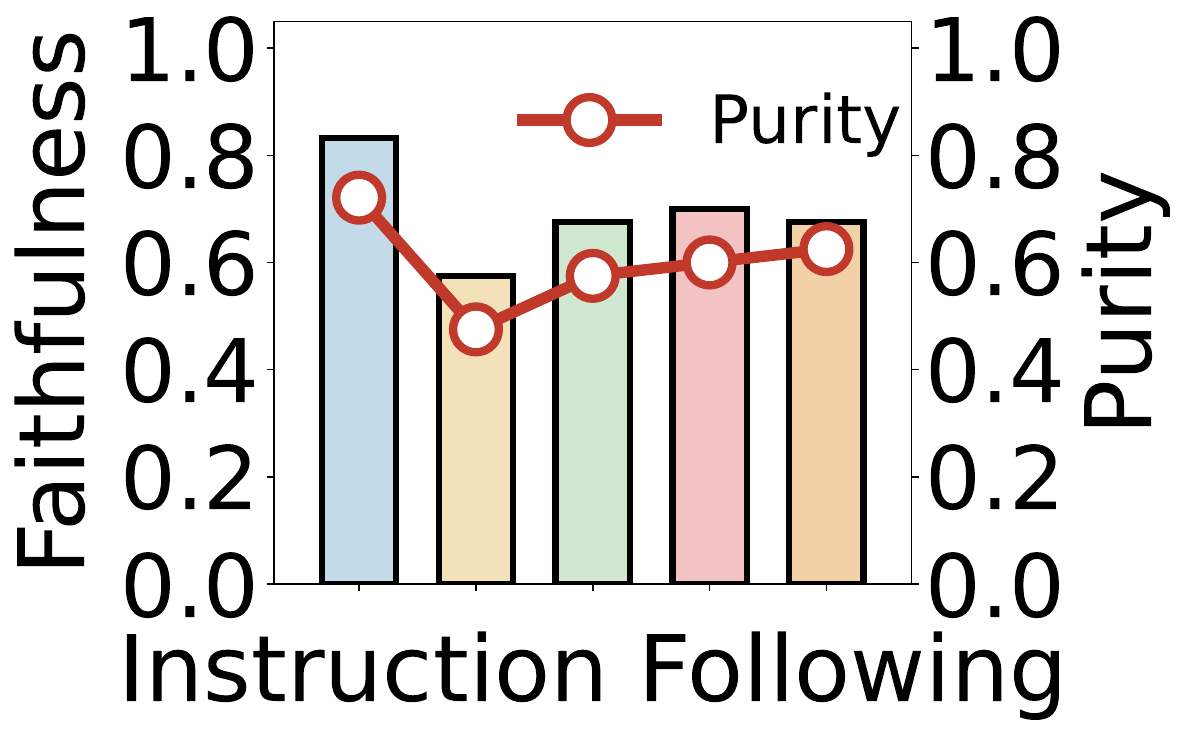}
  \end{minipage}
  \hfill
  \begin{minipage}[b]{0.32\linewidth}
    \centering
    \includegraphics[width=\linewidth]{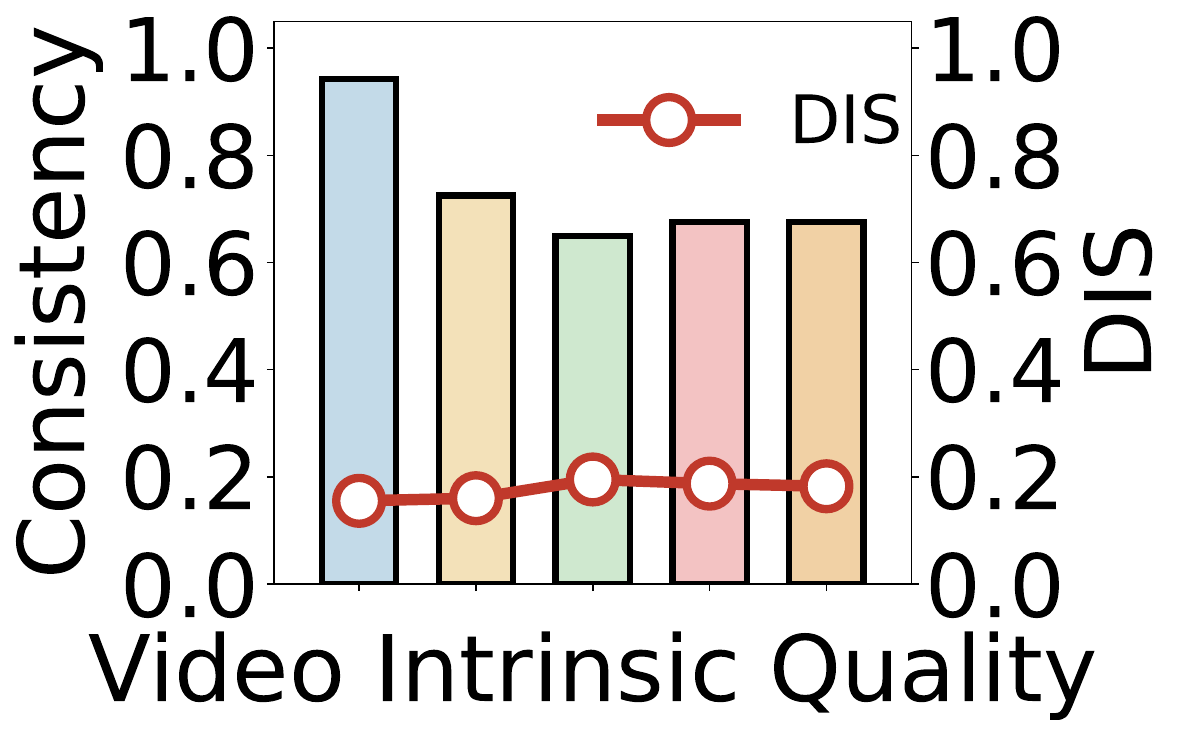}
  \end{minipage}
  \vspace{-0.1in}
  \caption{Tuning Performance of different ablations.}
  \label{fig:ablation_tuning}
  \vspace{-0.05in}
\end{figure}

We conduct ablations in Figure~\ref{fig:ablation_tuning} and Table~\ref{tab:ablation}. Removing auxiliary storyboard tasks mainly degrades coherence and temporal consistency, ablating the retrieval reward causes the largest coherence drop, and SFT-only preserves engagement but weakens faithfulness and purity, showing that multi-task supervision and GRPO-based visual grounding are both important for continuity and instruction adherence. For reward modeling and synthesis grounding, depth is the dominant retrieval cue: removing it nearly collapses retrieval, while removing pose also causes a large drop and contrastive-loss or hard-negative ablations are milder. For video generation, the full depth--pose design achieves the most balanced performance, confirming that spatial layout and character dynamics are both necessary for faithful and consistent short-drama synthesis. \vspace{-0.05in}

\begin{table}[t]
  \centering
  \small
  \setlength{\tabcolsep}{1.8mm}
  \renewcommand{\arraystretch}{1.1} 
  \caption{Retrieval performance comparison.}
  \label{tab:retrieval}
  \vspace{-0.12in}
  \begin{tabular}{c|cccc}
    \hline
    Method & R@10 & N@10 & R@20 & N@20 \\
    \hline
    tongyi-vision-flash & 0.0687 & 0.0395 & 0.1008 & 0.0475 \\
    tongyi-vision-plus & 0.0795 & 0.0455 & 0.1130 & 0.0540 \\
    \hline
    LightRAG & 0.0781 & 0.0451 & 0.1056 & 0.0521 \\
    RAGAnything & 0.0833 & \underline{0.0472} & 0.1176 & 0.0558 \\
    VidoRAG & \underline{0.0909} & 0.0469 & \underline{0.1241} & \underline{0.0597} \\
    \hline
    \model & \textbf{0.0954} & \textbf{0.0483} & \textbf{0.1429} & \textbf{0.0603} \\
    \hline
  \end{tabular}
  \vspace{-0.1in}
\end{table}

\subsection{Retrieval Discriminability (RQ3)}
We evaluate retrieval discriminability against general-purpose vision-language encoders and retrieval-augmented methods in Table~\ref{tab:retrieval}; \model\ achieves the best result, showing that bidirectional contrastive alignment with same-drama hard-negative mining yields stronger text-visual discrimination than generic multimodal pretraining. Even VidoRAG, the closest retrieval-trained competitor, trails \model\ on every metric, indicating that explicit depth--pose projection provides visual grounding beyond retrieval-oriented objectives alone. The advantage is largest at the larger recall cutoff while NDCG gaps remain narrow, suggesting that the reward model mainly broadens candidate coverage and that top-position precision remains difficult for all methods; overall, purpose-built text-visual alignment establishes a stronger retrieval foundation for downstream generation. \vspace{-0.05in}

\begin{figure}[t]
  \centering
  % --- Legend ---
  \includegraphics[width=\linewidth, trim={5 10 5 10}, clip]{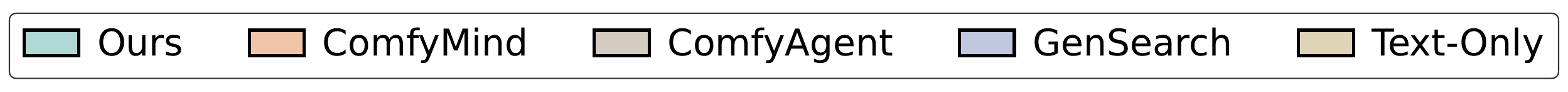}
  \vspace{-0.15in} 

  % --- Bar/line chart group ---
  \begin{subfigure}[b]{0.236\textwidth}
    \centering
    \includegraphics[width=\linewidth]{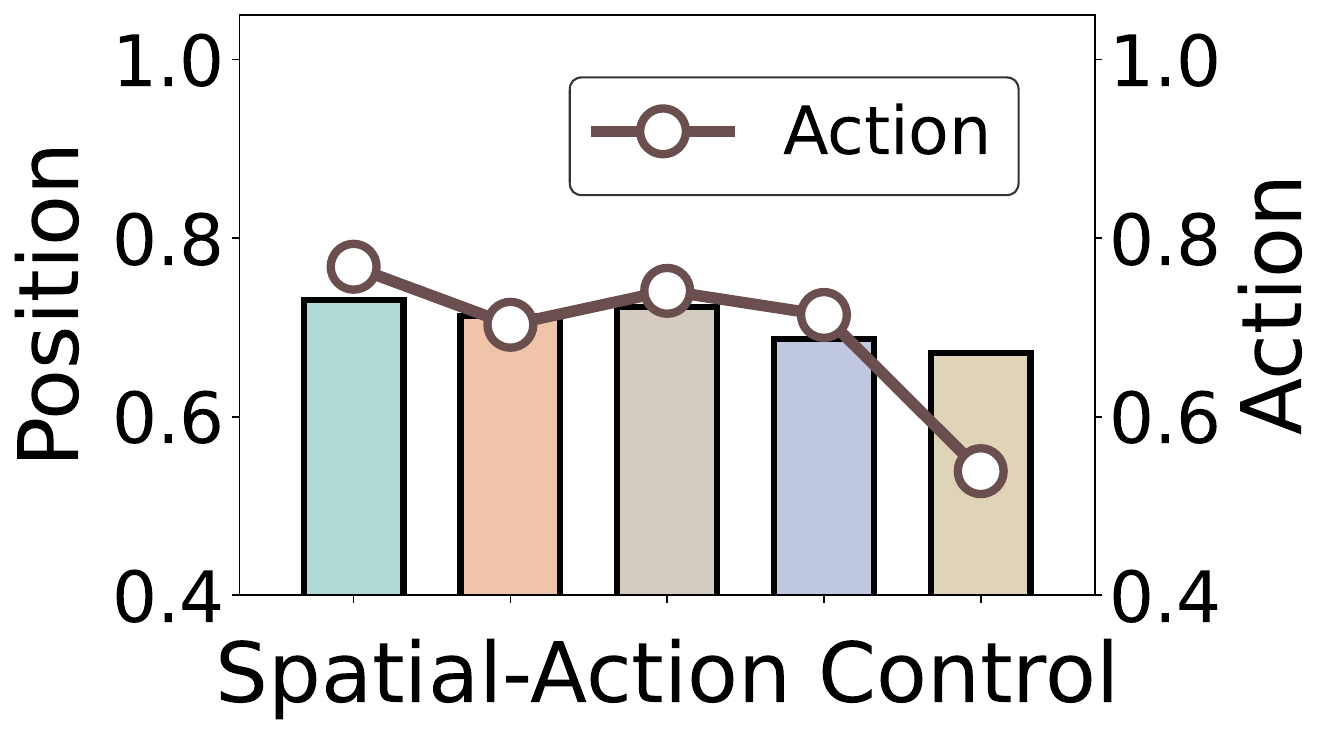}
    \label{fig:h2}
  \end{subfigure}
  \hfill
  \begin{subfigure}[b]{0.236\textwidth}
    \centering
    \includegraphics[width=\linewidth]{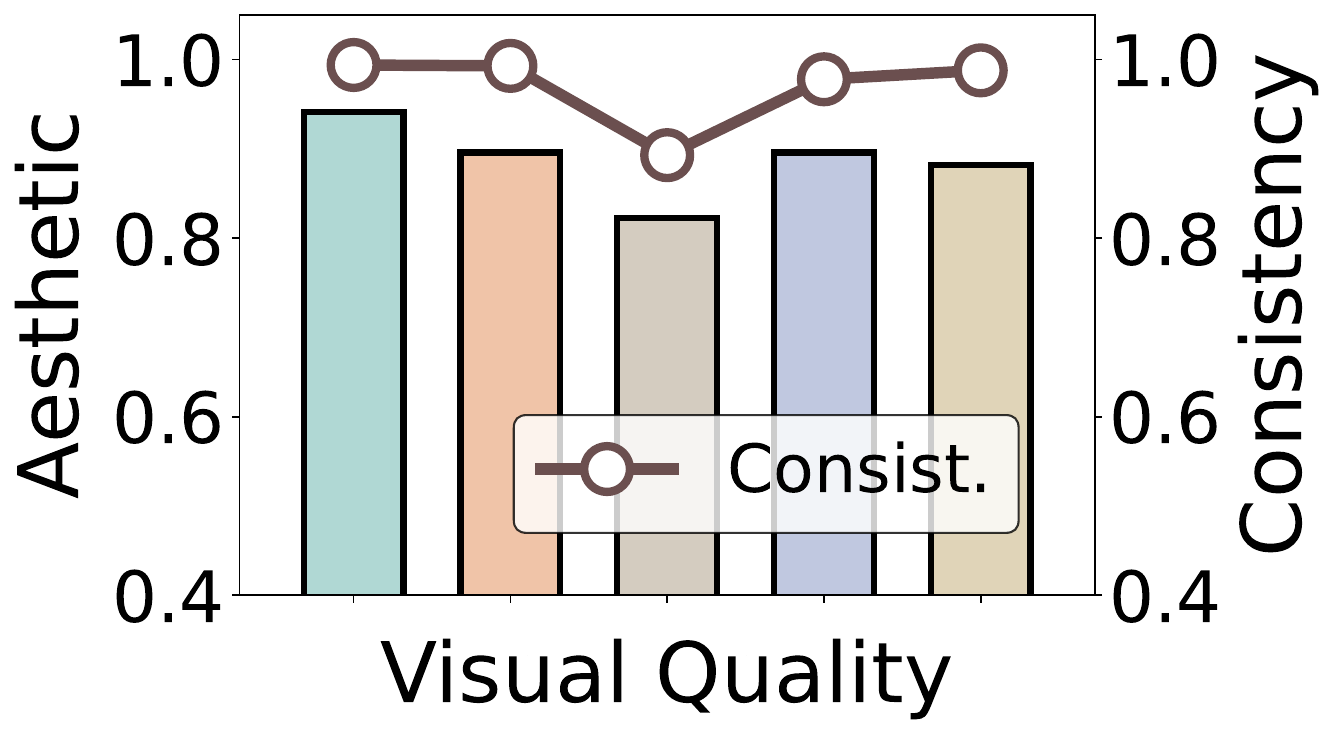}
    \label{fig:h1}
  \end{subfigure}
  
  \vspace{-0.3in} 
  \caption{Quality of First-Frame Generation.}
  \label{fig:hyper}
  \vspace{-0.05in} 
\end{figure}

\subsection{Quality of First-Frame Generation (RQ4)}

We evaluate retrieval-grounded first-frame generation against ComfyMind, ComfyAgent, GenSearch, and a Text-Only baseline, with results shown in Figure~\ref{fig:hyper}. \textbf{Visual Quality.} \model\ leads on Aesthetics and Consistency, while ComfyAgent exhibits the weakest aesthetics and the text-only baseline trails on both, confirming that retrieval grounding provides visual references critical for photorealism and cross-frame stability. \textbf{Spatial-Action Control.} \model\ leads on both Position and Action accuracy, with the largest margin on Action where text-only generation largely fails to capture character dynamics, indicating that pose-conditioned retrieval is essential for character action fidelity. \textbf{Synergy.} Consistent with the ablation in Table~\ref{tab:ablation}, depth and pose exhibit complementary roles---depth constraining spatial layout, pose governing character dynamics. \vspace{-0.05in}

\begin{figure*}[t]
  \centering
  \includegraphics[width=1\linewidth]{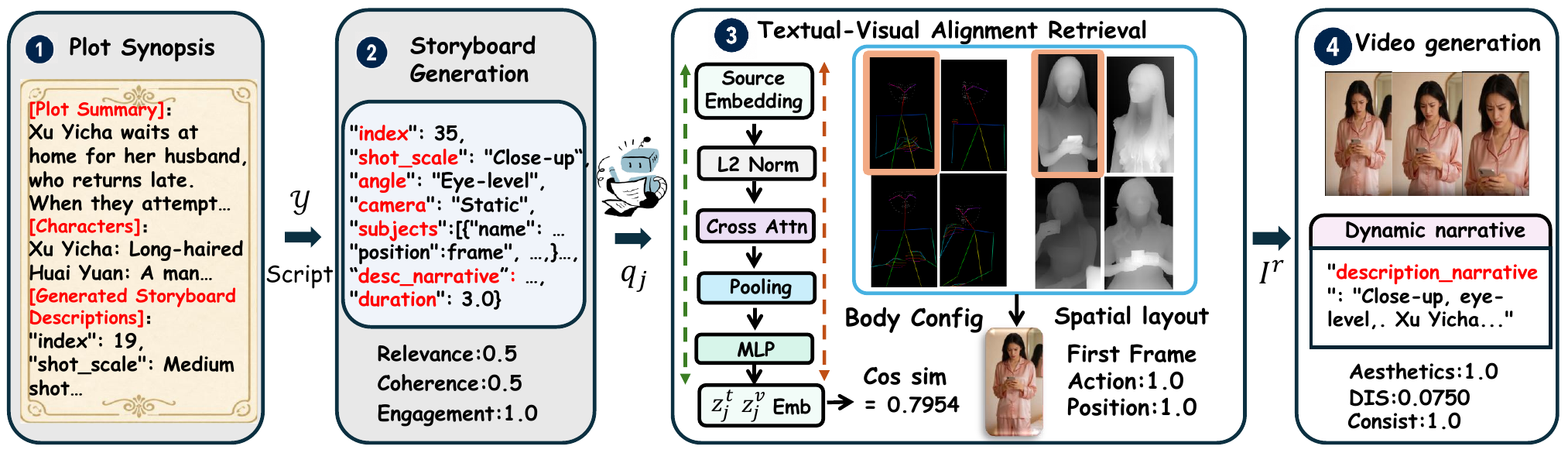}
  \vspace{-0.2in} 
  \caption{The end-to-end short drama generation process}
  \vspace{-0.15in}
  \label{fig:case_study}
\end{figure*}

\begin{table}[t]
\centering
\setlength{\tabcolsep}{1.5mm}
\renewcommand{\arraystretch}{1.05}
\caption{Hyperparameter Study on SFT steps}
\label{tab:sft_checkpoint_eval}
\small
\vspace{-0.12in}
\begin{tabular}{c|cc|cc}
\hline
\multirow{2}{*}{Step} & \multicolumn{2}{c|}{Narrative Quality} & \multicolumn{2}{c}{Intrinsic Quality} \\
\cline{2-5}
 & Relevance & Engagement & Consistency & DIS $\downarrow$\\
\hline
40 & 0.6050 & 0.4767 & 0.7200 & 0.1938 \\
120 & 0.6683 & 0.5284 & 0.7967 & 0.1740 \\
240 & 0.6367 & 0.5667 & 0.8867 & 0.1643 \\
360 & 0.7050 & 0.5867 & 0.8683 & 0.1581 \\
\hline
Ours & \textbf{0.7230} & \textbf{0.6032} & \textbf{0.9003} & \textbf{0.1549} \\
\hline
\end{tabular}%
\vspace{-0.1in}
\end{table}

\begin{figure}[t]
  \centering
  
  \begin{subfigure}[b]{0.48\linewidth}
    \centering
    \includegraphics[width=\linewidth, trim={7 5 10 5}, clip]{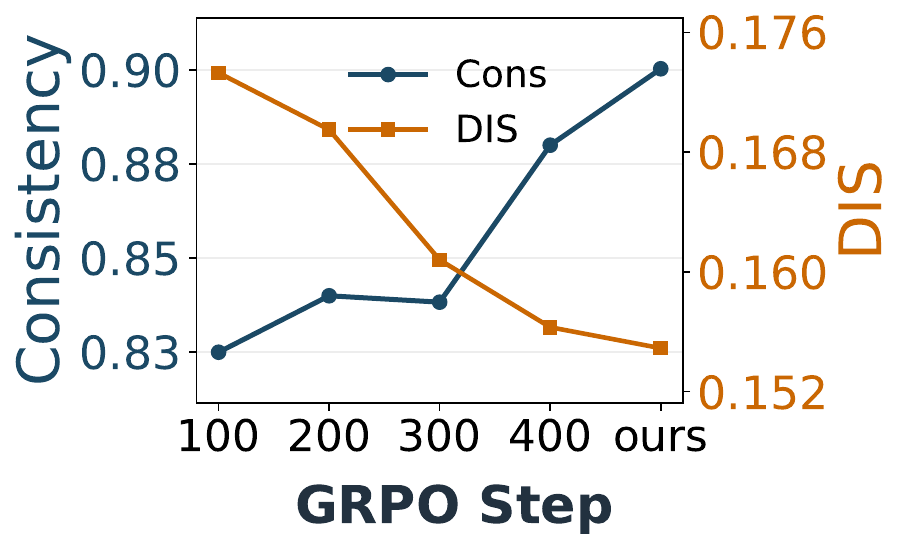}
    \vspace{-0.18in}
    \label{fig:grpo_video_quality}
  \end{subfigure}
  \hfill
  \begin{subfigure}[b]{0.48\linewidth}
    \centering
    \includegraphics[width=\linewidth, trim={7 5 10 5}, clip]{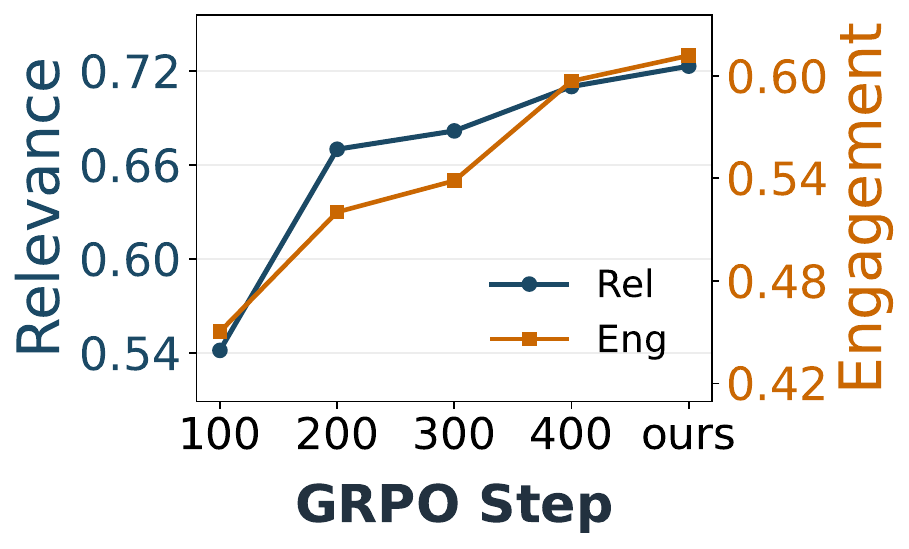}
    \vspace{-0.24in}
    \label{fig:grpo_narrative_quality}
  \end{subfigure}

  \caption{Hyperparameter study on GRPO steps}
  \vspace{-0.2in}
  \label{fig:grpo_checkpoint_eval}
\end{figure}

\subsection{Hyperparameter Study (RQ5)}
Table~\ref{tab:sft_checkpoint_eval} and Figure~\ref{fig:grpo_checkpoint_eval} study training-step hyperparameters for SFT and GRPO. \textbf{SFT steps.} Intermediate SFT checkpoints show non-monotonic behavior: narrative scores fluctuate, consistency generally improves, and DIS steadily decreases as training proceeds. The selected checkpoint at step~426 achieves the best results across all four metrics, indicating that sufficient schema supervision is needed to balance storyboard grounding and temporal stability. \textbf{GRPO steps.} The GRPO curves show the strongest overall balance around the selected checkpoint at step~441, with improved consistency and lower DIS while preserving competitive relevance and engagement, suggesting that our policy optimization further aligns storyboard planning with generation quality.

\subsection{Case Study (RQ6)}
To qualitatively demonstrate \model's effectiveness, Figure~\ref{fig:case_study} illustrates its end-to-end short drama generation process. Given an input of a plot summary, character profiles, and contextual storyboard history, it generates a structured storyboard script. In Step 2, the model outputs precise cinematic controls, such as 'Close-up', 'Eye-level', and 'Static' shots. Rather than relying on raw text, the Step 3 retrieves spatial layouts and body configurations aligned with the generated script. This retrieval-augmented grounding informs the video generation phase (Step 4), ensuring the final narrative faithfully reflects specified actions while maintaining visual aesthetics and temporal consistency.

\vspace{-0.2in}
\section{Related Works}
\label{sec:relate}

\noindent\textbf{Short-Drama Generation}.
Short-drama generation maps compact narrative inputs to dense shot sequences, shooting scripts, and dialogue-centered visual scenes. Recent work has begun to explore task-relevant formatting schemes and controllable generation frameworks for this direction ~\citep{shouldershot, dreamrunner}. SkyScript studies script--shooting-script structure for short dramas ~\citep{skyscript100m}, while SkyReels-A2 further studies controllable video synthesis by composing multiple visual elements in generated scenes~\citep{skyreels}. Existing generation frameworks still largely rely on prompt-level control, whereas \model\ internalizes short-drama production patterns and uses retrieved depth--pose references to mitigate the limitations of text-only generation.

\noindent\textbf{SFT and RL Post-training}.
Recent foundation-model research treats supervised fine-tuning and reinforcement learning as complementary stages for aligning generation behavior, with video-specific RL improving temporal reasoning and GRPO-style refinement advancing multimodal alignment~\citep{li2025video_rl_tuning,deepseekr1, visionr1}. In visual generation, Gen-Searcher, VisionCreator, and VisionCreator-R1 apply SFT or GRPO to multi-step visual creation~\citep{feng2026gen,visioncreator,visioncreatorr1}. Different from these systems, \model\ tailors post-training to short-drama storyboard generation, emphasizing structured shot planning and visual grounding.

\noindent\textbf{Multimodal RAG}.
 Multi-RAG integrates video, audio, and text streams for adaptive video understanding, while structured RAG frameworks capture long-range spatio-temporal dependencies in video evidence~\citep{arefeen2024vita, mao2025multi_rag,fu2026videostir}. Recent work further analyzes retrieval configurations, reranking, generation-time evidence integration, and retrieval over interleaved text-image documents aligned with downstream VLM preferences~\citep{mragdesign}. In contrast, \model\ does not retrieve generic documents, RGB images, or motion examples alone; it retrieves depth--pose references that encode short-drama spatial layout and character configuration, making the evidence directly actionable for first-frame generation.

\vspace{-0.05in}
\section{Conclusion}

This paper frames plot-to-short-drama generation as a visually grounded storyboard planning problem and proposes \model, which couples schema-constrained multi-shot planning with controllable video realization through multi-task supervision, a text-visual alignment reward, and retrieval-grounded depth--pose references. We further construct \dataset, a shot-level benchmark with aligned keyframes, dialogue, structured storyboards, and protocols spanning narrative, instruction-following, and video quality. Experiments show \model\ improves storyboard coherence, controllability, and generation quality over representative baselines. Future work will explore broader genres, tighter audio-visual synchronization, and interactive storyboard editing.

% This paper systematically studies plot-to-short-drama generation as a visually grounded storyboard planning problem. We propose \model, which connects schema-constrained multi-shot planning with controllable video realization through multi-task storyboard supervision, a text-visual alignment reward, retrieval-grounded depth--pose references, and multi-objective policy optimization. We further construct \dataset, a shot-level benchmark with aligned keyframes, dialogue, structured storyboards, and evaluation protocols spanning narrative quality, instruction following, and video quality. Experiments demonstrate that \model\ improves storyboard coherence, visual controllability, and generation quality over strong baselines. Future work will explore broader video genres, tighter audio-visual synchronization, and more interactive storyboard editing.

\section*{Limitations}
While \model\ improves visual grounding through retrieval-augmented depth--pose references, retrieval can still be imperfect for highly complex or long-tail interactions, especially when the reference gallery lacks scenes with comparable spatial layouts or body configurations. We mitigate this issue during first-frame generation by explicitly framing the retrieved conditions as reference cues rather than hard constraints: pose maps are used to indicate action and body configuration, while depth maps are used to indicate spatial layout and scene geometry. The generation prompt emphasizes that, when the retrieved reference conflicts substantially with the target storyboard text, the textual constraints should take precedence. 
\section*{Ethical Considerations}
The \dataset\ benchmark is constructed from 35 live-action short dramas for which we hold the necessary copyrights, usage rights, and distribution permissions. All video frames, transcripts, and structured annotations used in this work are derived from these legally authorized sources, so the dataset does not involve unauthorized use of third-party copyrighted content. The released dataset is intended solely for academic research and is not licensed for commercial use. For the human evaluation, participants voluntarily assessed anonymized and randomly shuffled outputs, and responses were collected anonymously for academic research only.

\bibliography{ref}
\appendix
\clearpage
\section{Appendix}
\subsection{Baseline Methods}

This appendix outlines the backbones, baselines, and format-based adaptations used for end-to-end short drama generation. We employ Vidu Q3 Turbo and Seedream v5 Lite as our primary video and image backbones, alongside Wan 2.6 and NanoBanana 2 to evaluate cross-generator generalizability. To demonstrate our framework's effectiveness, we evaluate it against four comparison categories: (1) short-drama formatting and generation methods, (2) image generation frameworks, (3) generative agentic frameworks, and (4) retrieval-augmented frameworks.

\vspace{0.1in}
\textbf{Short-Drama Generation Methods}
\setlength{\leftmargini}{10pt}
\begin{itemize}

  \item {\textbf{SkyScript}}~\citep{skyscript100m}: We instantiate this comparison by following its script--shooting-script format: the input plot is serialized into a shooting-script-style shot plan, and the resulting shot descriptions are passed to the same downstream image and video generation interface used by the other generation baselines.
  
  \item {\textbf{ShoulderShot}}~\citep{shouldershot}: Generates over-the-shoulder dialogue videos using dual-shot generation and looping mechanisms for spatial and character consistency.
  
  \item {\textbf{SkyReels-A2}}~\citep{skyreels}: An elements-to-video (E2V) framework assembling visual elements into scenes via image-text joint embedding for global coherence.
  
  \item {\textbf{Dreamrunner}}~\citep{dreamrunner}: Combines LLM layout planning, retrieval-augmented motion adaptation, and spatial-temporal control for story-to-video synthesis.

\end{itemize}

\textbf{Image Generation Frameworks}
\setlength{\leftmargini}{10pt}
\begin{itemize}
  \item {\textbf{ComfyMind}}~\citep{guo2026comfymind}: A collaborative AI system for general-purpose generation, featuring a Semantic Workflow Interface and search tree planning with localized feedback.
  
  \item {\textbf{Gen-Search}}~\citep{feng2026gen}: A search-augmented agent that utilizes multi-hop reasoning to search online for textual knowledge and visual references for grounded image generation.

  \item {\textbf{ComfyAgent}}~\citep{comfybench}: An LLM-based framework that autonomously designs collaborative multi-agent AI systems by generating executable workflows.
\end{itemize}

\textbf{Generative Agentic Frameworks}
\setlength{\leftmargini}{10pt}
\begin{itemize}

  \item {\textbf{MovieAgent}}~\citep{movieagent}: Employs hierarchical CoT to organize scripts, scenes, and cinematography for long-form video generation.

  \item {\textbf{UniVA}}~\citep{univa}: Decomposes user intentions into modular steps for versatile video understanding, editing, and generation.

  \item {\textbf{VideoAuteur}}~\citep{videoauteur}: Synthesizes long narrative videos by planning captions, actions, and visual states for coherent generation.
  
 \item {\textbf{GENMAC}}~\citep{genmac}: Iteratively refines complex text-to-video prompts through multi-agent design, generation, and redesign.
\end{itemize}

\textbf{Retrieval-Augmented Frameworks}
\setlength{\leftmargini}{10pt}
\begin{itemize}
  \item {\textbf{Tongyi-Embedding-Vision}}~\citep{alibaba_tongyi_vision}: A multimodal embedding API from Alibaba Cloud that projects images and text into a unified vector space for efficient cross-modal retrieval and semantic alignment.
  \item {\textbf{LightRAG}}~\citep{lightrag}: Combines structural indexing and dual-level graph retrieval for efficient knowledge extraction.

  \item {\textbf{RAG-Anything}}~\citep{raganything}: Enables versatile multimodal retrieval (e.g., text, images, tables) via cross-modal graph reasoning.

  \item {\textbf{VidoRAG}}~\citep{pathorag}: A multimodal RAG framework featuring joint text-image search and multi-turn reasoning.
  
\end{itemize}
\begin{figure*}[!htbp]
  \centering
  \vspace*{-0.90in}
  \makebox[\textwidth][c]{
    \includegraphics[
      height=1\textheight,
      width=1.35\textwidth,
      keepaspectratio
    ]{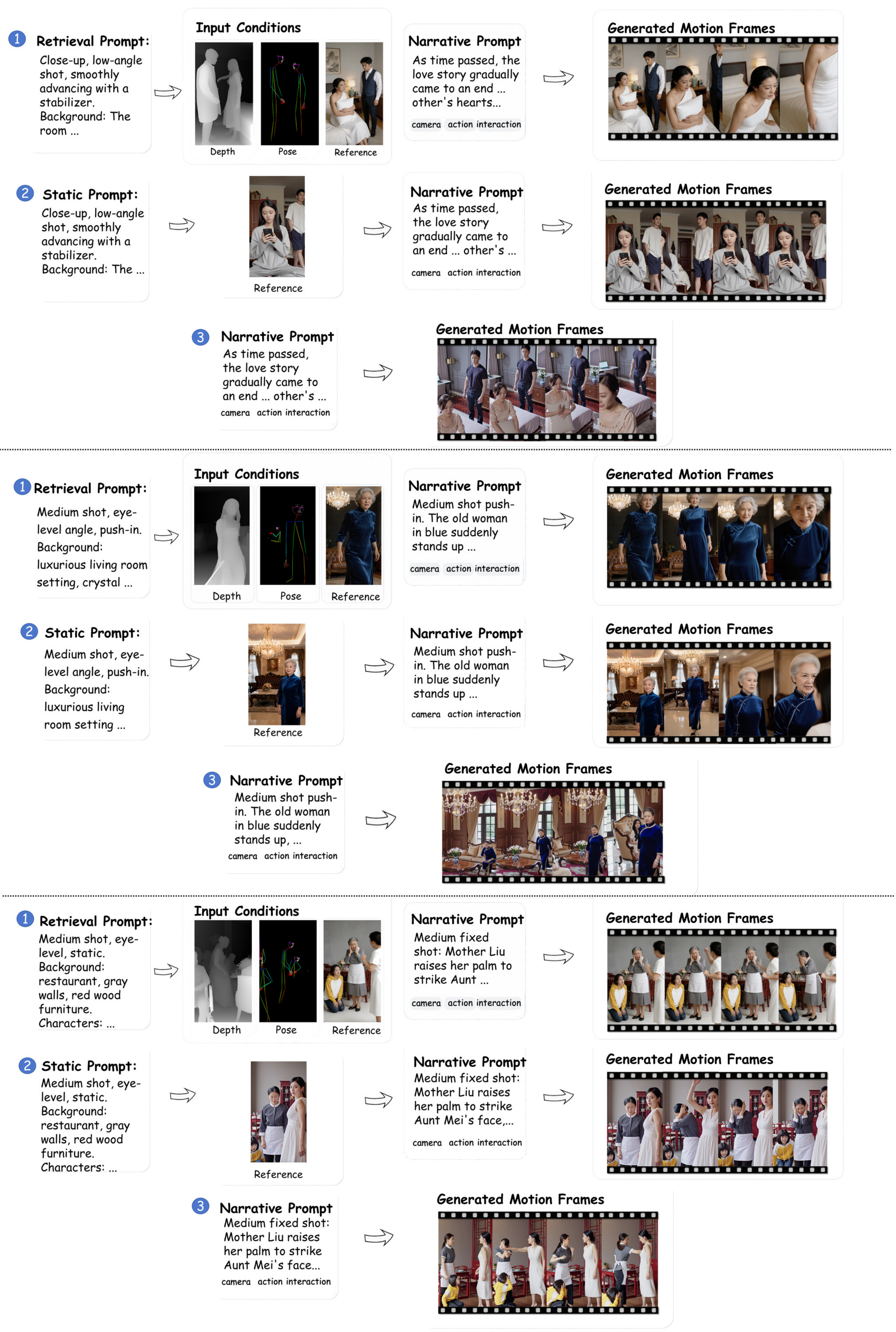}
  }
  \vspace{-0.35in}
  % Caption aligned with the surrounding discussion.
  \caption{Comparison of generation workflows between \model, image-to-video, and text-to-video }
  \label{fig:Comparison of generation workflows}
  \vspace{-0.20in}
\end{figure*}

\subsection{Generation Workflow Comparison}
To capture the cinematic language of real short dramas, we design a retrieval-augmented workflow between storyboard text and video generation. Figure~\ref{fig:Comparison of generation workflows} shows three representative shots, each implemented with three workflows: our method, image-to-video, and text-to-video. Compared with text-to-video, our method reduces the semantic gap between storyboard text and visual generation by retrieving depth and skeleton conditions from a short-drama-native storyboard library, leading to better control over shot composition, spatial layout, and character motion. Compared with image-to-video, our method uses depth and skeleton as cleaner structural priors, preserving scene hierarchy and implicit shot-level cues while avoiding the appearance noise in raw RGB reference images. As a result, our workflow produces results that are more controllable, consistent, stable, and better aligned with short-drama storyboard design.

\subsection{Hyperparameter Study}

% --- Table placed directly below the subsection heading ---
\begin{table}[!ht]
\centering
\small
\setlength{\tabcolsep}{2.2mm}
\renewcommand{\arraystretch}{1.05} 
\caption{Study on depth and pose fusion weights.}
\label{tab:depth-pose-weight-sweep}
\vspace{-8pt} 
\begin{tabular}{c|c|r|r|r|r} 
\hline
\multicolumn{2}{c|}{Weights} & \multicolumn{1}{c|}{R@10}  & \multicolumn{1}{c|}{N@10}  & \multicolumn{1}{c|}{R@20} & \multicolumn{1}{c}{N@20} \\
\hline
\multirow{5}{*}{\rotatebox{90}{\textbf{Depth/Pose}}} 
& 0.1/0.9 & 7.12 & 3.45 & 11.58 & 4.57 \\
\cline{2-6} 
& 0.2/0.8 & 8.43 & 4.20 & 14.04 & 5.62 \\
\cline{2-6}
& 0.3/0.7 & 4.51 & 2.11 & 7.85 & 2.95 \\
\cline{2-6}
& 0.4/0.6 & \textbf{9.74} & \textbf{5.07} & \textbf{14.48} & \textbf{6.26} \\
\cline{2-6}
& \textbf{0.5/0.5}* & 9.54 & 4.83 & 14.29 & 6.03 \\
\hline
\end{tabular}%
\end{table}

% --- Figure placed after the table ---
\begin{figure}[!ht]
  \centering
  
  \begin{subfigure}[b]{0.48\linewidth}
    \centering
    \includegraphics[width=\linewidth]{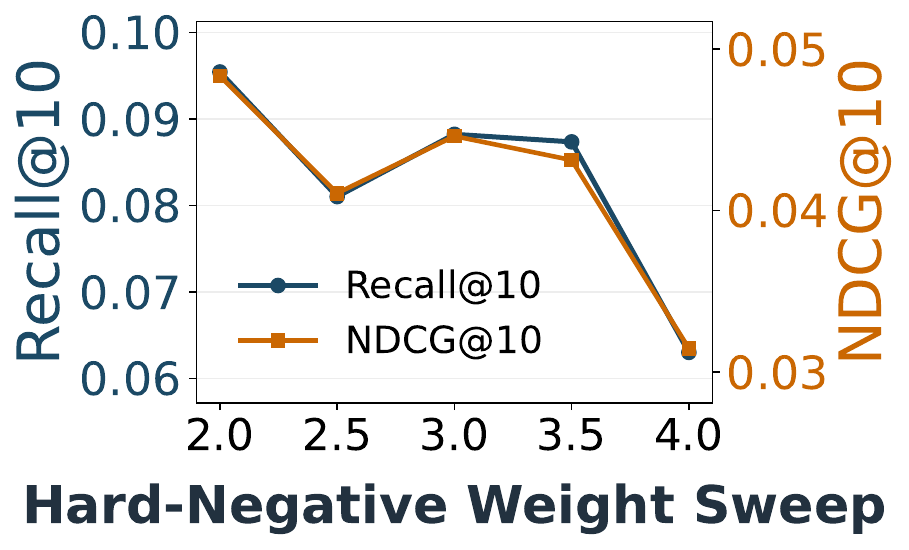}
    \vspace{-0.18in}
    \label{fig:hard_lambda}
  \end{subfigure}
  \hfill
  \begin{subfigure}[b]{0.48\linewidth}
    \centering
    \includegraphics[width=\linewidth]{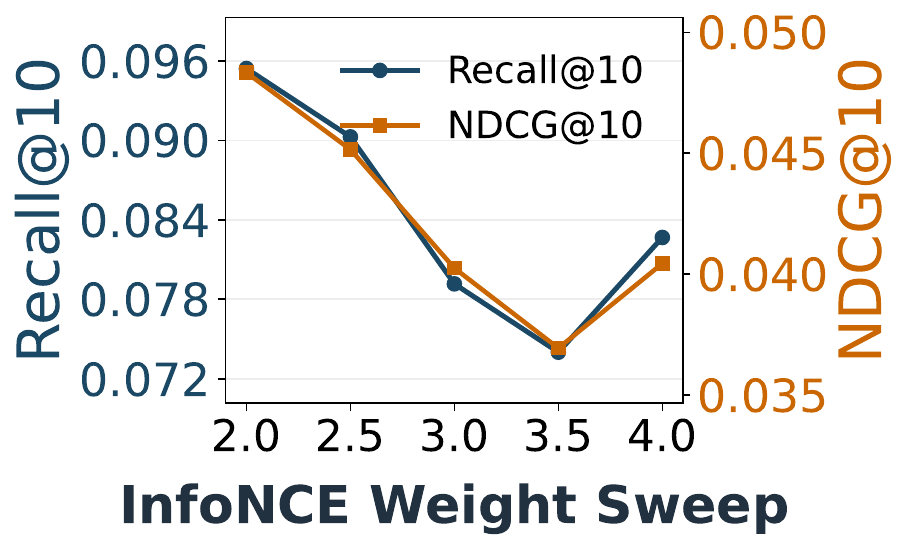}
    \vspace{-0.18in}
    % \caption{InfoNCE Lambda} 
    \label{fig:infonce_lambda}
  \end{subfigure}
  % \vspace{-0.25in}
  \caption{Balancing Negative and Hard Negative Losses} 
  \label{fig:contrastive_ablation}
\end{figure}

% --- Text immediately follows the figure ---
We examine two hyperparameter groups that govern retrieval alignment. \textbf{Depth--pose fusion weights} (Table~\ref{tab:depth-pose-weight-sweep}) show that extreme weighting degrades retrieval. While a slightly pose-emphasized setting (0.4/0.6) achieves the best overall retrieval metrics, its performance gain over equal fusion (0.5/0.5) is marginal (e.g., only a 0.24 difference in N@10). During downstream video synthesis, we empirically observed that decreasing the depth weight below 0.5 occasionally compromises the strict preservation of scene geometry and camera layout. Therefore, we adopt equal weights (0.5/0.5) to balance multi-person pose configurations with stable spatial structures in the generated videos. \textbf{Contrastive loss weights} (Figure~\ref{fig:contrastive_ablation}) show that standard and hard-negative losses are both necessary, respectively supporting global feature separation and fine-grained discrimination; over-weighting either objective harms retrieval, indicating that generic inter-sample negatives and same-drama hard negatives provide complementary rather than redundant supervision.

\subsection{Benchmarking}
\label{sec:benchmark}

We construct the benchmark from raw short-drama videos through a staged pipeline, producing a shot-level storyboard corpus for both supervised training and downstream generation. Unless otherwise specified, the multimodal analysis, transcript correction, and storyboard annotation stages use Qwen3.5-Plus as the annotation backbone.

\paragraph{Shot preprocessing}
Each episode is segmented into shots with one representative keyframe. A boundary at time $t$ is kept if
\begin{equation}
d_t > \tau,\qquad \Delta_t \ge L_{\min},
\end{equation}
where $d_t$ denotes the inter-frame visual difference score, $\Delta_t$ is the resulting segment length, $\tau=30.0$ is the cut threshold, and $L_{\min}=15$ frames is the minimum shot length. We retain shot indices, timestamps, durations, keyframe paths, and source-video paths, which are reused in later alignment and annotation stages.

\paragraph{Depth--pose preprocessing}
For each representative keyframe, we extract the geometric controls used by retrieval and first-frame conditioning. Depth maps are produced with Depth Anything V2~\citep{depth_anything_v2}, using the ViT-L encoder checkpoint and an input size of 518, then min--max normalized to an 8-bit control image. Pose maps are produced with DWPose~\citep{dwpose}, using the ONNX whole-body pipeline with YOLOX-L human detection and the \texttt{dw-ll\_ucoco\_384} pose estimator. The rendered pose control includes body, hand, and face keypoints on a blank canvas, so the downstream gallery stores geometry rather than appearance.

\paragraph{Gallery construction and split policy}
The depth--pose gallery is constructed from the available short-drama episodes and is used as a contextual visual prior rather than as a target storyboard label. To prevent episode-level leakage during evaluation, we remove all shots from the source episode of each test sample before retrieval. Thus, generated storyboards cannot retrieve depth--pose references from the same episode as the held-out target, and the held-out storyboard annotations are used only for evaluation. Retrieval instead operates over depth and pose references from other episodes, providing reusable cinematographic geometry while avoiding memorization of episode-specific layouts. This setting preserves the practical benefit of a short-drama geometry gallery while enforcing a stricter separation between test targets and retrieval evidence.

\paragraph{Episode-level semantic annotation}
We next run a full-video multimodal analysis pass to extract a character list, a plot summary, and a global dialogue reference. These episode-level conditions provide cross-shot semantic grounding for later annotation and off-screen speaker attribution, while also improving consistency in character naming and narrative interpretation across the episode.

\begin{figure}[ht]
  \centering
  \includegraphics[width=\linewidth, trim={0 155 0 155}, clip]{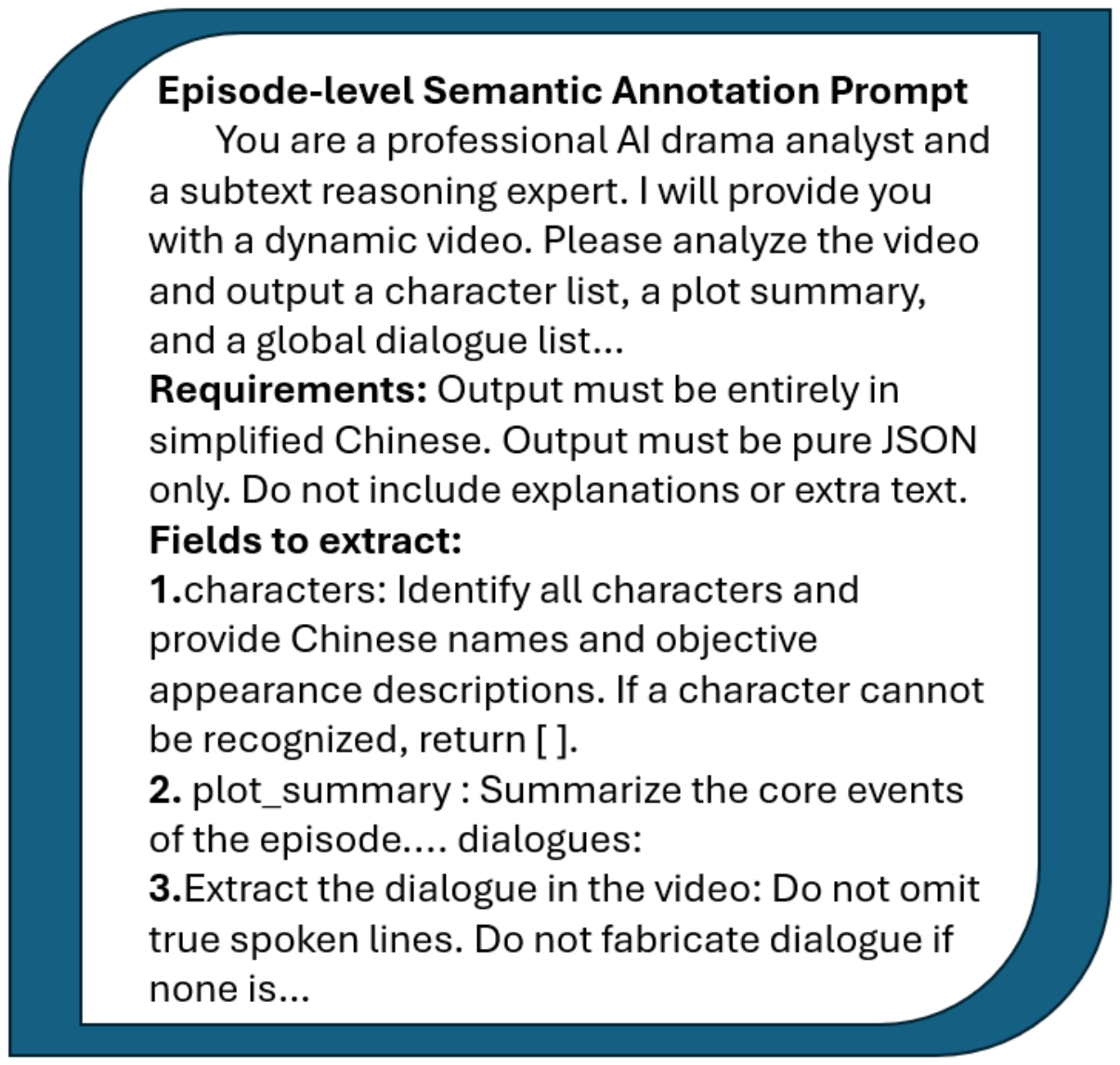}
  \vspace{-0.28in}
  \caption{Episode-level semantic annotation prompt}
  \label{fig:semantic_annotation_prompt}
\end{figure}

\paragraph{Shot-level transcript correction}
We align speech to shot boundaries, run ASR on each shot, and apply an LLM-based correction step conditioned on the character list, plot summary, and dialogue reference. This step fixes recognition errors while preserving shot indices and empty transcripts. It is particularly useful for short-drama speech with colloquial expressions, aliases, omitted subjects, and off-screen utterances.

\begin{figure}[ht]
  \centering
  \includegraphics[width=\linewidth, trim={0 180 0 160}, clip]{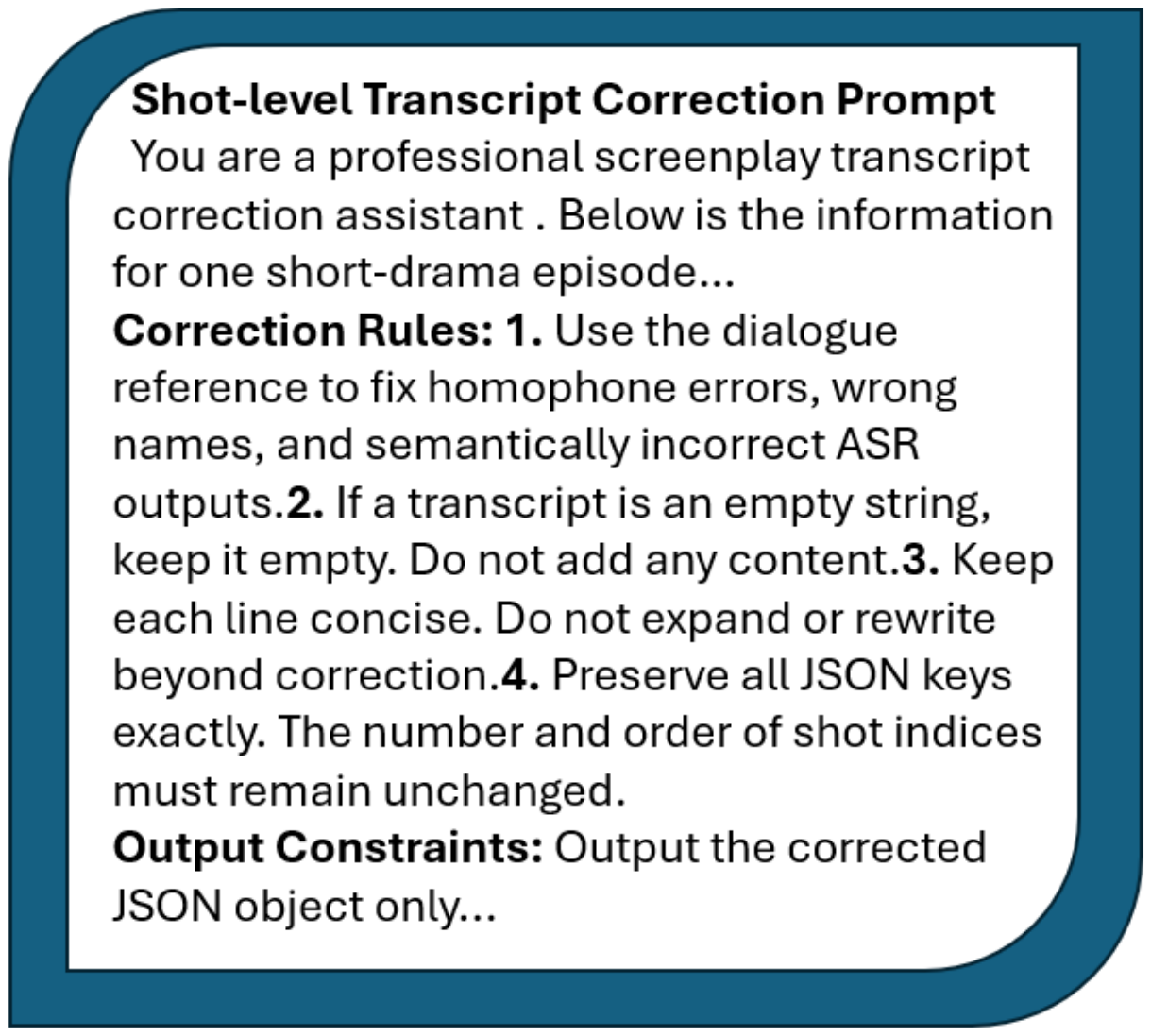}
  \vspace{-0.25in}
  \caption{Shot-level transcript correction prompt}
  \label{fig:transcript_correction_prompt}
\end{figure}

\begin{figure*}[!ht] 
  \centering
  \vspace{-0.8in} 
  
  \includegraphics[width=\textwidth, trim={100 80 100 80}, clip]{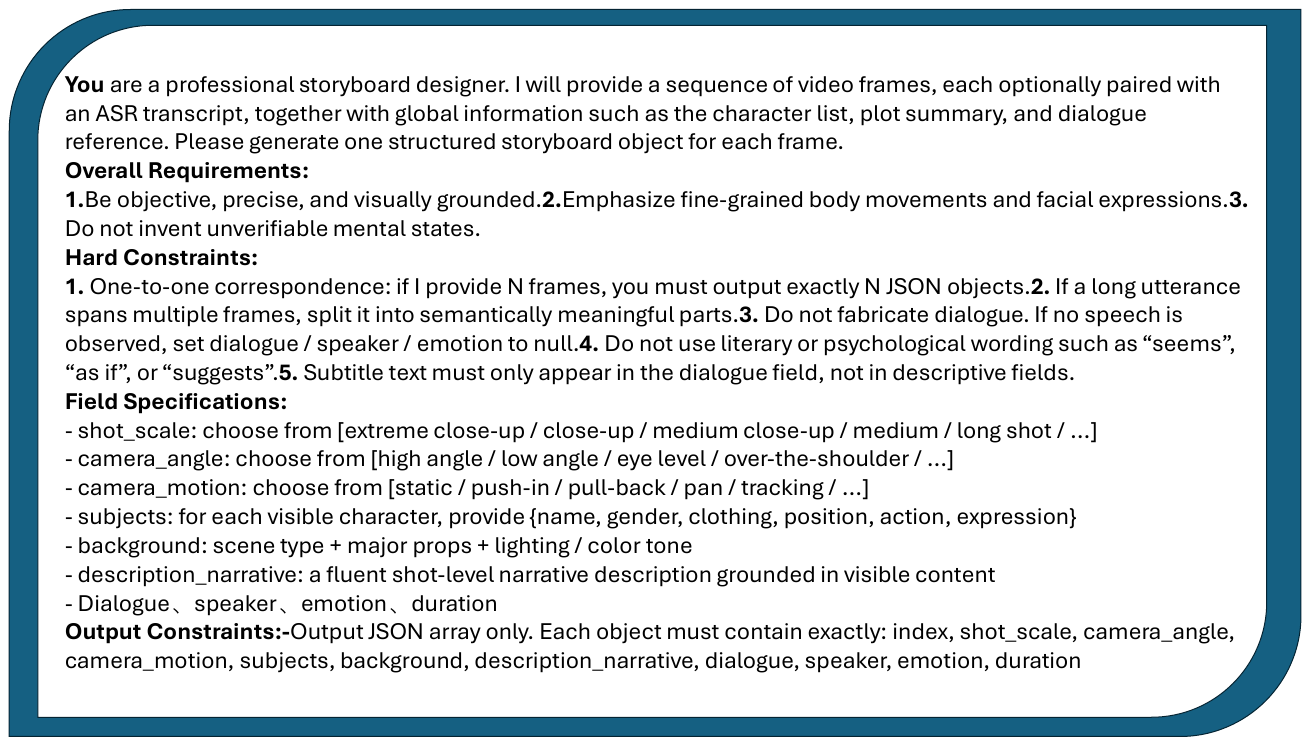}
  \vspace{-0.5in} 
  \caption{Shot-level storyboard annotation prompt}
  \label{fig:Storyboard_Annotation_Prompt}
  
  \vspace{0.5in} 
  
  \includegraphics[width=\textwidth, trim={100 80 100 80}, clip]{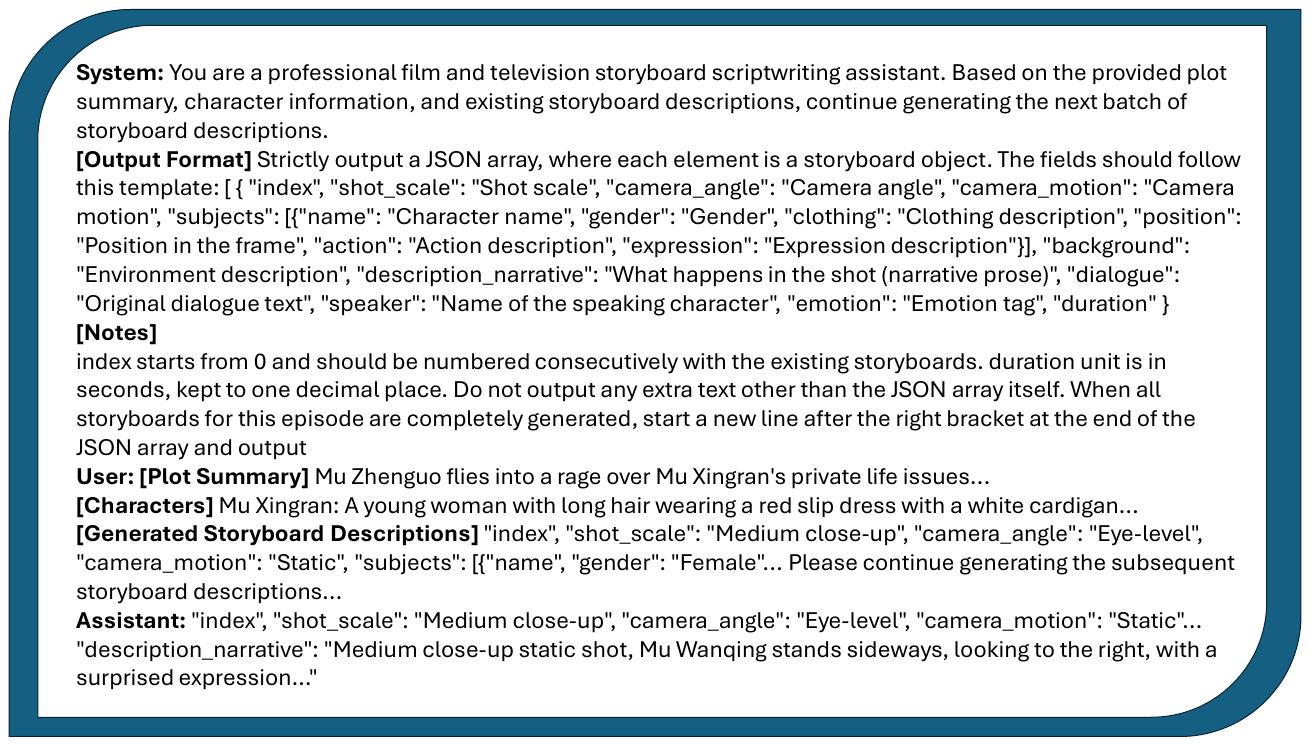}
  \vspace{-0.5in} 
  \caption{An example SFT prompt for storyboard continuation. Long contexts are truncated for readability}
  
  \label{fig:sft_prompt_example}
  
  \vspace{-0.3in} 
\end{figure*}

\paragraph{Shot-level storyboard annotation}
Given the keyframe, shot duration, corrected transcript, and episode-level context, we prompt a multimodal model to produce one structured storyboard object per shot, as shown in \ref{fig:Storyboard_Annotation_Prompt}. Each object includes camera fields, subjects, background, narrative, dialogue, speaker, emotion, and duration. The prompt enforces one-to-one frame-to-shot correspondence, avoids unverifiable psychological descriptions, and resolves off-screen speech to the speaker. This stage converts multimodal signals into a consistent representation suitable for retrieval and generation.

% \paragraph{Multi-task SFT benchmark}
% Storyboard sequences are organized into a multi-task SFT benchmark centered on continuation, alongside tasks for single-shot prediction, field completion, shot ordering, and summary-guided reconstruction. These tasks provide supervision over local continuity, structured completion, temporal coherence, and cross-shot semantic consistency in long-form generation. The continuation subset is stored as \texttt{system}/\texttt{user}/\texttt{assistant} triplets, while all tasks are serialized with a chat template before training.

% \paragraph{TTS usage}
% In the application pipeline, the \texttt{dialogue}, \texttt{speaker}, and \texttt{emotion} fields are further consumed by a TTS module to synthesize aligned speech tracks for generated episodes with speaker-aware expressive voice control. This makes the structured storyboard not only a supervision target, but also a practical intermediate representation for end-to-end short-drama production.

\paragraph{Multi-task SFT}
We organize storyboard sequences into a multi-task SFT centered on continuation, with one example shown in Figure~\ref{fig:sft_prompt_example}. Auxiliary tasks include single-shot prediction, field completion, shot ordering, and summary-guided reconstruction, providing supervision for local continuity, structured completion, temporal coherence, and cross-shot semantic consistency. The continuation subset is stored as \texttt{system}/\texttt{user}/\texttt{assistant} triplets, and all tasks are serialized with a chat template for training.

\paragraph{TTS usage}
In the application pipeline, the \texttt{dialogue}, \texttt{speaker}, and \texttt{emotion} fields are consumed by a TTS module to synthesize speaker-aware expressive speech. Thus, the structured storyboard serves both as a supervision target and as an intermediate representation for end-to-end short-drama production.

\subsection{Multimodal Judge Prompt}
\label{sec:judge_prompt}

\begin{figure}[!ht]
  \centering
  
  \includegraphics[width=\textwidth, trim={100 120 100 110}, clip]{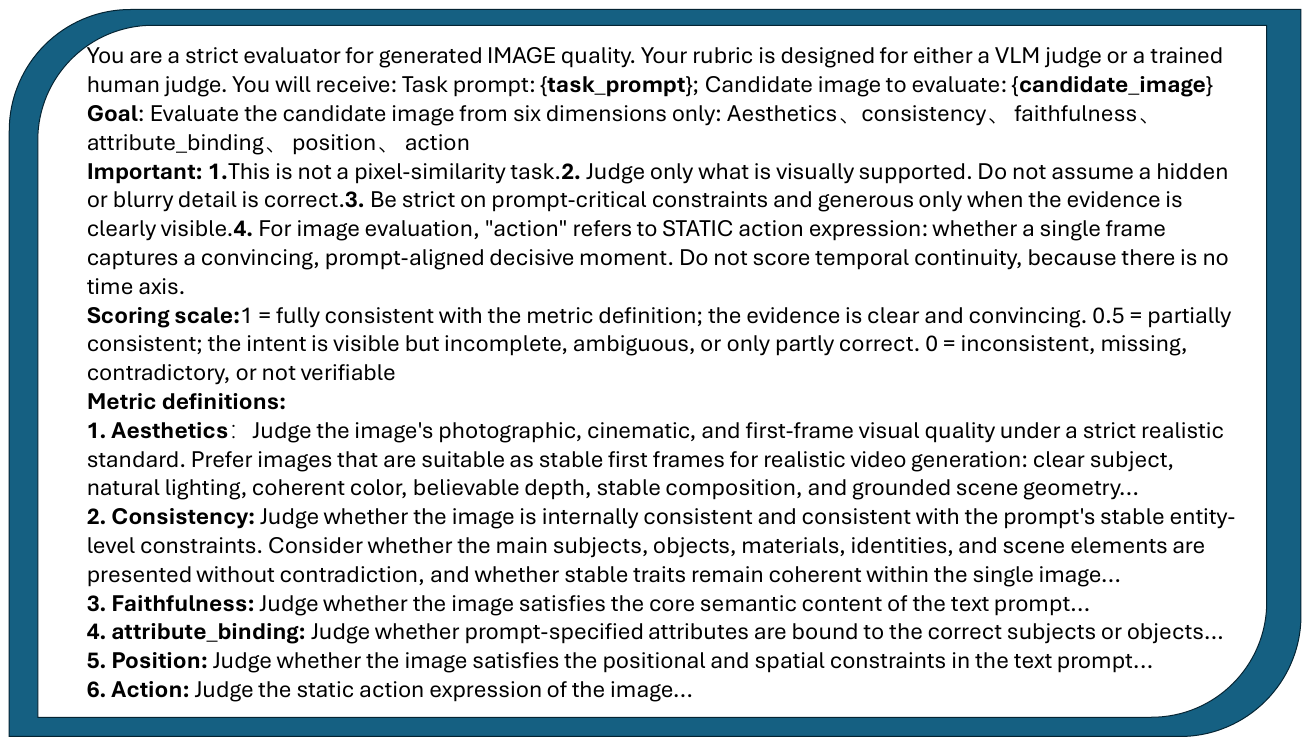}
  \vspace{-0.1in}
  \caption{Detailed evaluation prompt and rubric used for assessing static image generation quality.}
  \label{fig:Image_Judge_prompt}

  \vspace{0.15in} 
  
  \includegraphics[width=\textwidth, trim={100 100 70 0}, clip]{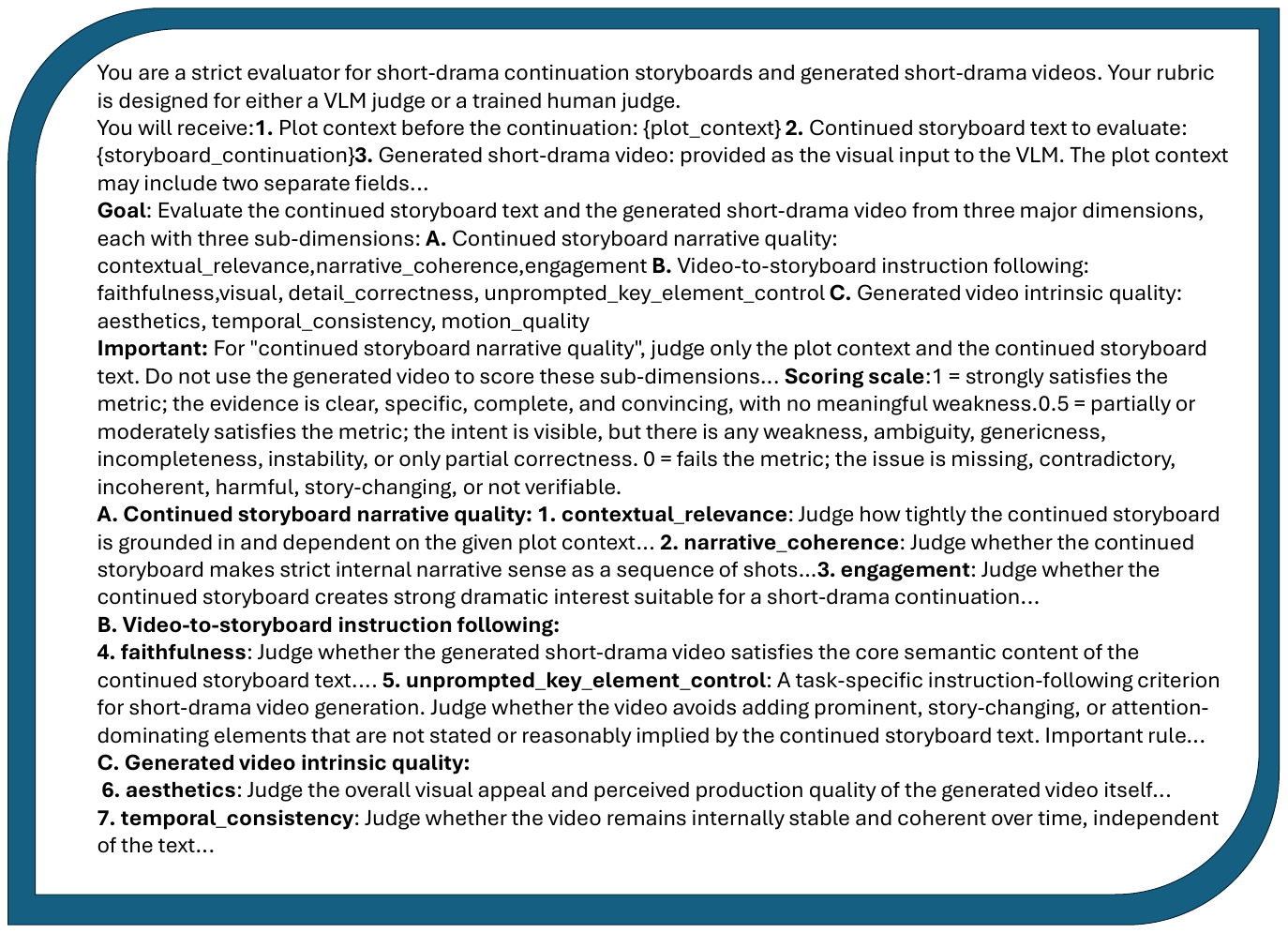}
  \vspace{-0.1in}
  \caption{Detailed evaluation prompt and rubric used for assessing temporal and dynamic video generation quality.}
  \label{fig:Video_Judge_prompt}
\end{figure}

To rigorously evaluate artifacts, we combine decomposed judge prompts (Figures~\ref{fig:Image_Judge_prompt} and \ref{fig:Video_Judge_prompt}) with automatic Similarity and DIS measurements. Judge prompts require metric-wise scores with evidence-grounded rationales, avoiding the ambiguity of holistic ratings and better capturing the multidimensional nature of short-drama generation, where quality depends on visual fidelity, prompt grounding, spatial layout, temporal execution, and narrative continuity.

For \textbf{Image Quality Assessment}, we evaluate each image along six dimensions: \textit{aesthetics}, \textit{consistency}, \textit{faithfulness}, \textit{attribute binding}, \textit{position}, and \textit{action}. This separation prevents visually appealing but semantically incorrect images from receiving inflated scores. Specifically, \textit{faithfulness} measures core semantic alignment, \textit{attribute binding} and \textit{position} verify correct modifier assignment and spatial relations, and \textit{action} assesses whether the frame captures a plausible decisive moment.

For \textbf{Video Quality}, we evaluate three axes. \textbf{Storyboard Narrative Quality} (\textit{Relevance}, \textit{Coherence}, \textit{Engagement}) assesses text-only continuation, so narrative flaws cannot be hidden by appealing visuals. \textbf{Instruction Following} (\textit{Faithfulness}, \textit{Similarity}, \textit{Purity}) verifies whether the video realizes the storyboard and dynamic narrative prompt while avoiding unprompted content. \textbf{Intrinsic Quality} (\textit{Aesthetics}, \textit{Consistency}, \textit{DIS}) measures visual fidelity and temporal stability independent of textual alignment; DIS is the masked SSIM distance after DIS-flow warping, where lower values indicate smoother adjacent-frame transitions.

All judge-scored metrics use a $\{0, 0.5, 1\}$ scale, where $1$ is reserved for complete and unambiguous satisfaction. Ambiguous or occluded details receive no credit, and character identity and story continuity are treated as hard constraints.

\vspace{-0.03in}
\subsection{Backbone Interfaces}
\label{app:generation_backbones}
\model\ uses commercial generation backbones only through explicit input--output interfaces. We do not claim novelty in these backbones; the reproducible component of our framework is how structured storyboards, retrieved depth--pose references, and dynamic narrative fields are converted into conditioning inputs for them.

\paragraph{Embedding extraction}
For retrieval and reward modeling, we use a frozen Tongyi-Embedding-Vision encoder to embed three inputs: the static visual condition $q_j$, the retrieved depth map, and the retrieved pose map. Text inputs are submitted as text records, while depth and pose controls are submitted as base64-encoded images. The resulting vectors are saved as per-shot arrays and then consumed by our trainable projection, fusion, contrastive alignment, and GRPO reward modules. Thus, Tongyi provides only fixed base embeddings; retrieval scoring and policy optimization are performed by our own alignment model.

\begin{figure}[ht]
  \centering
  \includegraphics[width=\linewidth, trim={260 80 140 15}, clip]{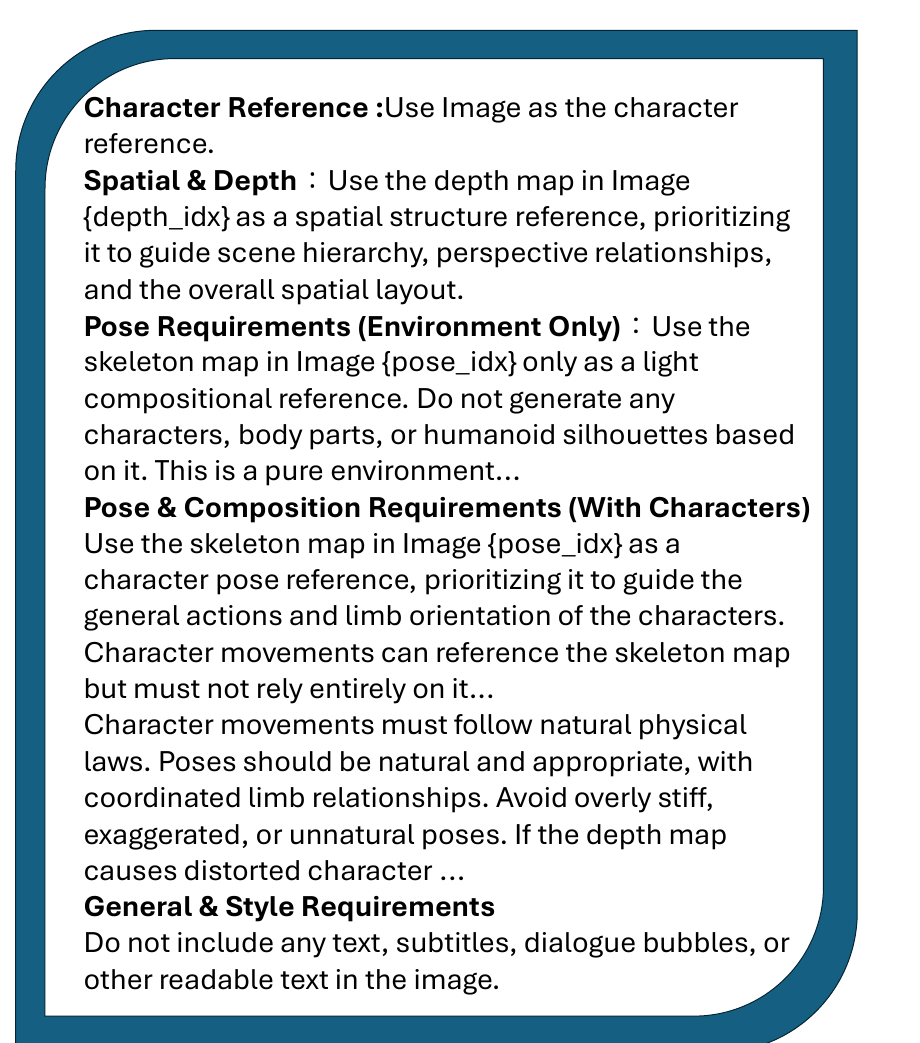}
  \vspace{-0.30in}
  \caption{Conditioning instructions appended to the storyboard-derived first-frame prompt.}
  \label{fig:post_frame_prompt}
\end{figure}

\paragraph{First-frame generation}
The first frame is not generated from the conditioning prompt in Figure~\ref{fig:post_frame_prompt} alone. For each shot, we first serialize the storyboard fields, including camera scale, camera angle, camera motion, subjects, background, and narrative description, into a shot-level prompt. Dialogue content is removed from this visual prompt because speech is handled by the dynamic narrative condition. We then append explicit conditioning instructions that bind the uploaded images to their roles: character reference images, if any, appear first; the retrieved depth map is uploaded next and is used for scene hierarchy, perspective, and spatial layout; the retrieved pose map is uploaded last and is used for coarse body configuration and limb orientation. The shot first frame is generated with the Seedream-v5-lite image-to-image interface at a vertical 9:16 resolution. If a depth or pose reference conflicts with the storyboard text, the storyboard text has the highest priority; for environment-only shots, the pose map is treated only as a weak compositional cue and must not introduce characters.

\paragraph{Image-to-video generation}
After first-frame synthesis, each shot is animated with the Vidu-q3-turbo image-to-video interface. The submitted payload contains the generated first frame, the dynamic narrative prompt derived from $d_j$, the shot duration, the target resolution, and the audio flag. In our experiments, video generation uses the first frame as the sole visual anchor and keeps audio generation disabled, so the visual comparison isolates storyboard-to-video realization rather than TTS quality.

\paragraph{Execution records}
For reproducibility, each API call stores the exact submitted prompt, uploaded control-image order, local paths of depth and pose references, selected character references, duration, resolution, task identifier, returned URL, and downloaded output path. These records make the commercial-backbone stage auditable even when the underlying service implementation is closed-source.

\begin{figure*}[t] 
  \centering
  \includegraphics[width=\textwidth, trim={0 210 0 210}, clip]{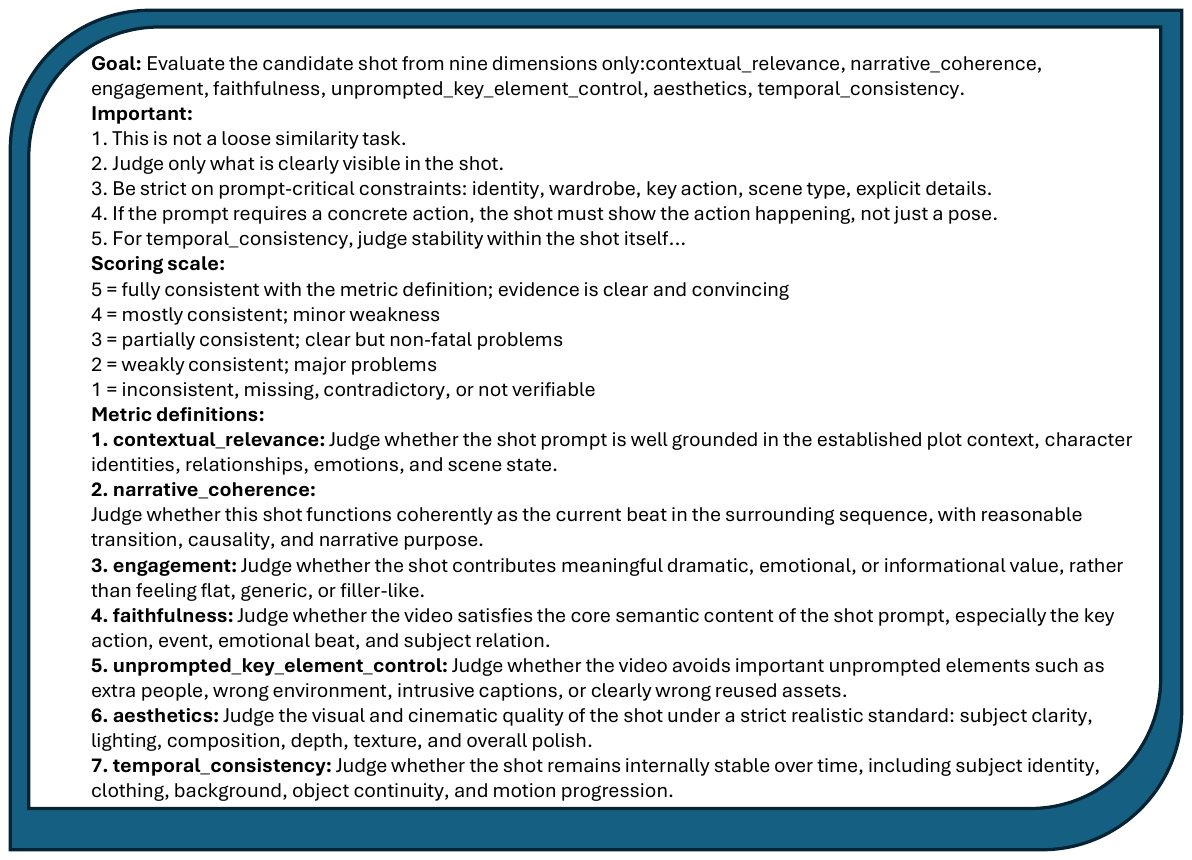}
  % \vspace{-0.3in} 
  \caption{Shot-Level Human Evaluation Rubric}
  \label{fig:human_prompt_example}
  % \vspace{-0.9in} 
\end{figure*}

\subsection{Human Evaluation}
\label{sec:human_eval}
\begin{figure}[t]
  \centering
  
  % --- 1. Legend ---
  \includegraphics[width=\linewidth]{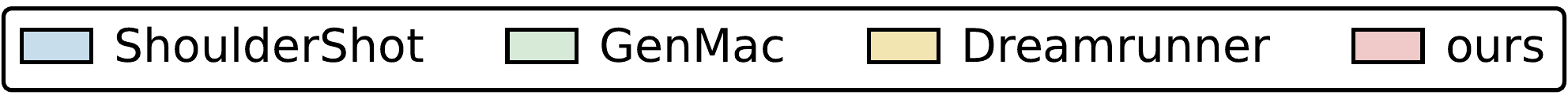}
  \vspace{-0.15in}

  % --- 2. Human Evaluation Metrics ---
  \begin{minipage}[b]{0.32\linewidth}
    \centering
    \includegraphics[width=\linewidth]{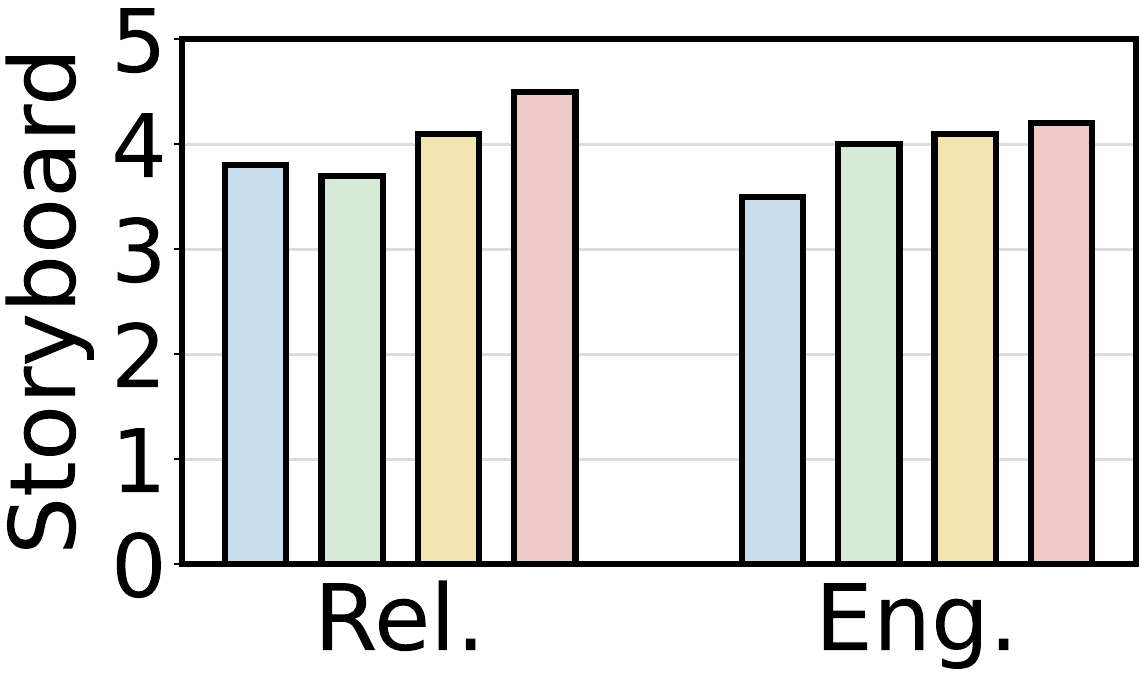}
  \end{minipage}
  \hfill
  \begin{minipage}[b]{0.32\linewidth}
    \centering
    \includegraphics[width=\linewidth]{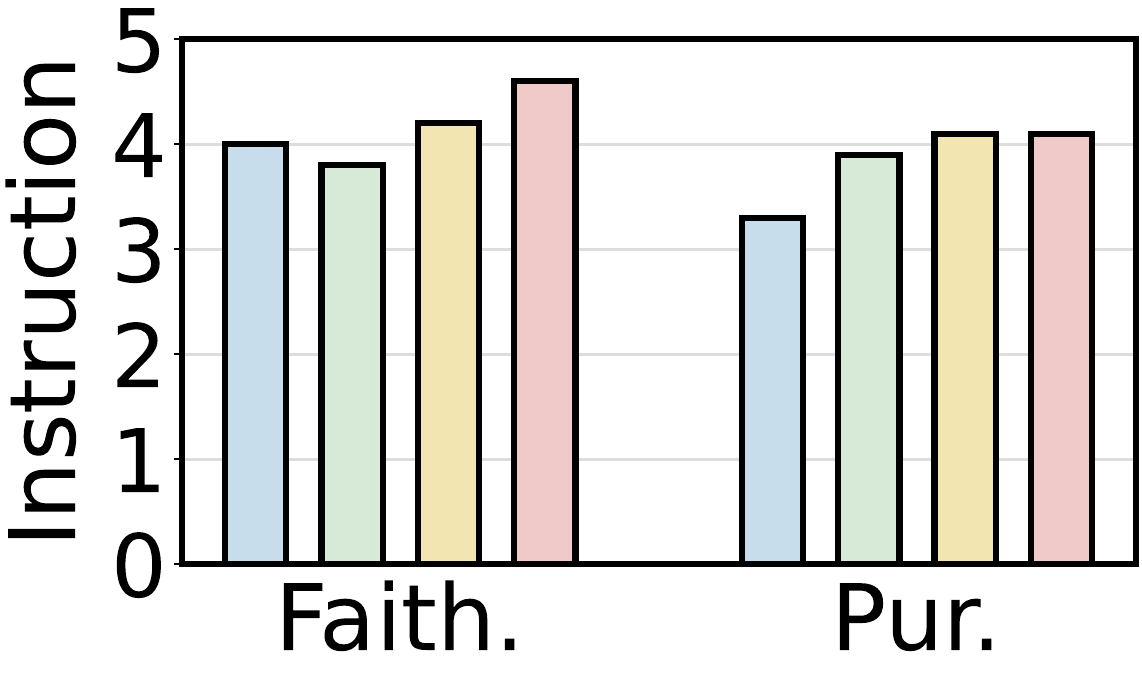}
  \end{minipage}
  \hfill
  \begin{minipage}[b]{0.32\linewidth}
    \centering
    \includegraphics[width=\linewidth]{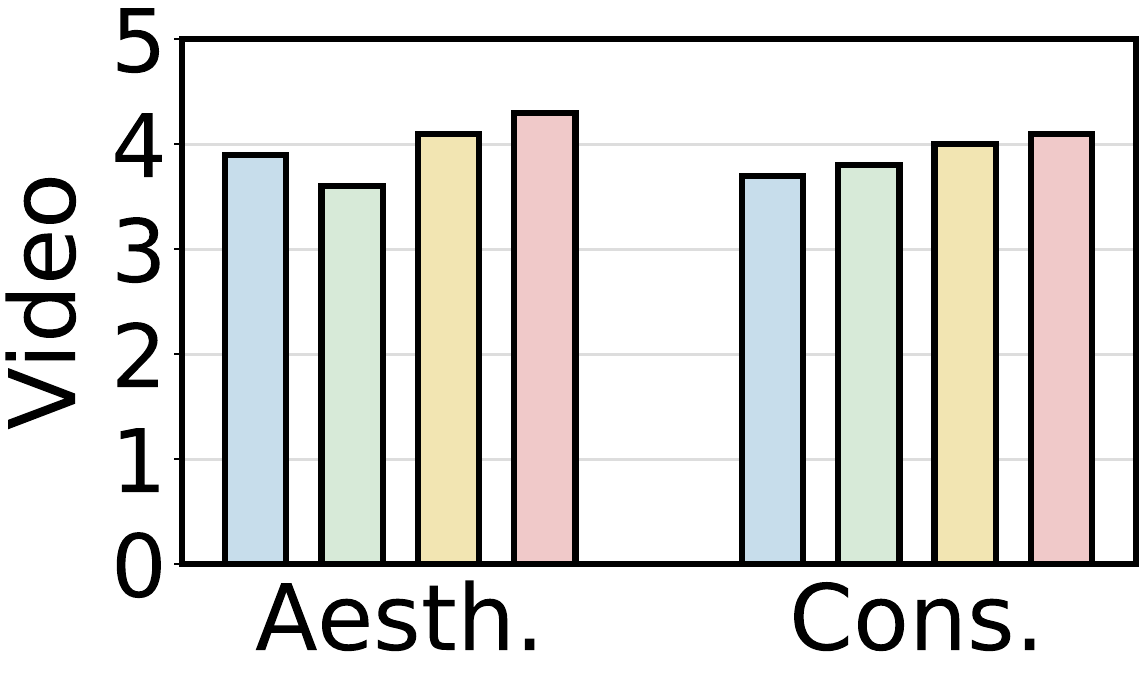}
  \end{minipage}
  \vspace{-0.1in}
  \caption{Human evaluation results across six metrics.}
  \label{fig:human_eval}
  \vspace{-0.05in}
\end{figure}

To complement the automatic judge-based evaluation, we further conduct a human study to directly assess the perceptual quality of generated short-drama videos. We randomly sample 30 generated video cases from the evaluation set, where each case contains 4--5 storyboard shots. For \model\ and three representative baselines, ShoulderShot, GenMac, and Dreamrunner, we recruit 23 student volunteers to evaluate anonymized and randomly shuffled outputs using a 5-point Likert scale. The study focuses on six human-judged dimensions: \textit{Relevance} and \textit{Engagement} for storyboard narrative quality, \textit{Faithfulness} and \textit{Purity} for instruction following, and \textit{Aesthetics} and \textit{Consistency} for intrinsic video quality. Scores from 1 to 5 indicate very poor, poor, fair, good, and excellent quality, respectively. The evaluation takes approximately 1.2 hours to complete, and all responses are collected anonymously and used only for academic research. As shown in Figure~\ref{fig:human_eval}, \model\ achieves the strongest or competitive human ratings across the six metrics, with clear advantages in narrative relevance, instruction faithfulness, visual aesthetics, and temporal consistency, indicating that geometry-guided planning and retrieval-grounded realization improve the perceived quality of short-drama generation.

\subsection{Computational Efficiency Discussion}
\model\ operates with a lightweight 8B planner during inference, resulting in computation costs comparable to standard small-scale LLMs. This allows storyboard generation without the high latency and API dependency associated with larger closed-source planners. Furthermore, the retrieval of static visual conditions relies on pre-computed depth and pose embeddings, adding limited overhead to the overall pipeline. Consequently, our framework supports controllable, end-to-end short drama generation while maintaining a modest computational footprint.
% \begin{figure*}[!ht] 
%   \centering
%   \vspace{-0.8in} 
%   \includegraphics[width=\textwidth, trim={0 0 0 0}, clip]{figs/judge.pdf}
%   \vspace{-0.3in} 
%   \caption{Shot-Level Human Evaluation Rubric}
  
%   \label{fig:human_prompt_example}
%   \vspace{-0.9in} 
% \end{figure*}

\clearpage 
\onecolumn 

\end{document}